\documentclass{article}
\usepackage{iclr2027_conference,times}

\usepackage{amsmath,amsfonts,bm}

\def\eqref#1{equation~\ref{#1}}

\def\1{\bm{1}}

\DeclareMathAlphabet{\mathsfit}{\encodingdefault}{\sfdefault}{m}{sl}
\SetMathAlphabet{\mathsfit}{bold}{\encodingdefault}{\sfdefault}{bx}{n}

\usepackage{hyperref}
\usepackage{xurl}
\usepackage{graphicx}
\usepackage{amsmath,amssymb,amsthm}
\usepackage{booktabs}
\usepackage{subcaption}
\usepackage{tikz}
\usetikzlibrary{positioning,arrows.meta,calc,fit,backgrounds,decorations.pathreplacing}
\usepackage[ruled,vlined]{algorithm2e}
\usepackage{xcolor}
\usepackage[T1]{fontenc}

\graphicspath{{figs/}}
\newlength{\tzs}\newlength{\tzg}   

\newcommand{\Tp}{T^{\dagger}}                 

\theoremstyle{plain}

\theoremstyle{definition}

\arxivcopy

\title{\hbox{Moore, Escher, Penrose: A Conformal Golden Braid}}

\author{
    \begin{tabular}[t]{c}
    \textbf{Sophia Feldman}\\
    Technion - Israel Institute of Technology\\
    \texttt{sophia.d@campus.technion.ac.il}
    \end{tabular}
    \And
    \begin{tabular}[t]{c}
    \textbf{Assaf Shocher}\\
    Technion - Israel Institute of Technology\\
    \texttt{assaf.sh@technion.ac.il}
    \end{tabular}
}

\begin{document}

\maketitle
\begin{figure}[h!]
  \vspace{-0.35cm}
  \hspace{1
  cm}\makebox[\textwidth][c]{
    \vbox{ 
      \setlength{\tzs}{0.185\textwidth}\setlength{\tzg}{0.01\textwidth}
      
      \hbox{
        \begin{minipage}[c]{\dimexpr 2\tzs+1\tzg\relax}
          \setlength{\fboxsep}{-3pt}
           \setlength{\fboxrule}{3pt}
          \fbox{\includegraphics[ width=\textwidth]{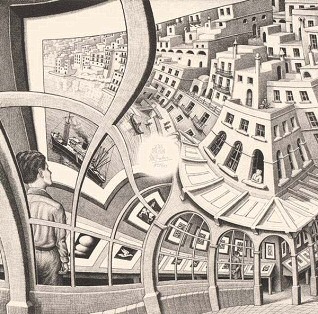}}\\[\tzg]
          \includegraphics[ width=\tzs]{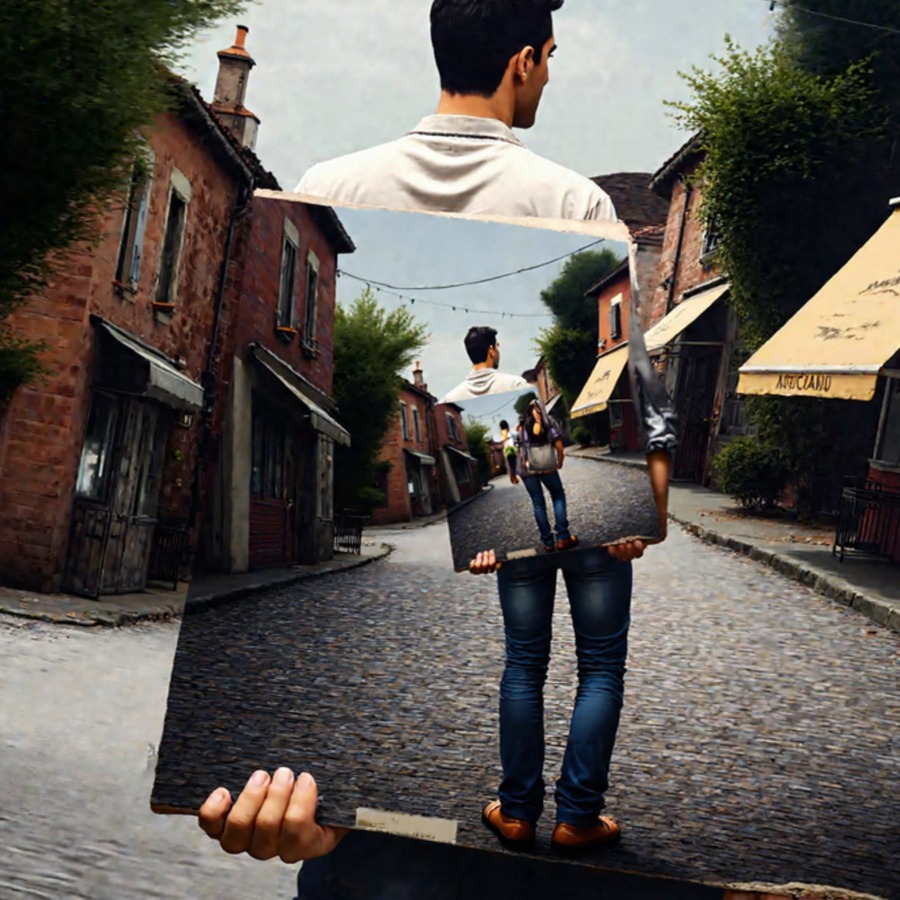}\hspace{\tzg}%
          \includegraphics[ width=\tzs]{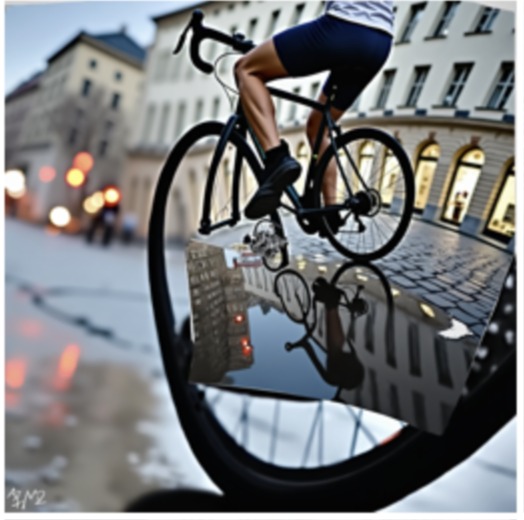}

        \end{minipage}%
        \hspace{\tzg}%
        \begin{minipage}[c]{\dimexpr 1\tzs-0.17\tzg \relax}
        \centering
          \includegraphics[ width=\textwidth]{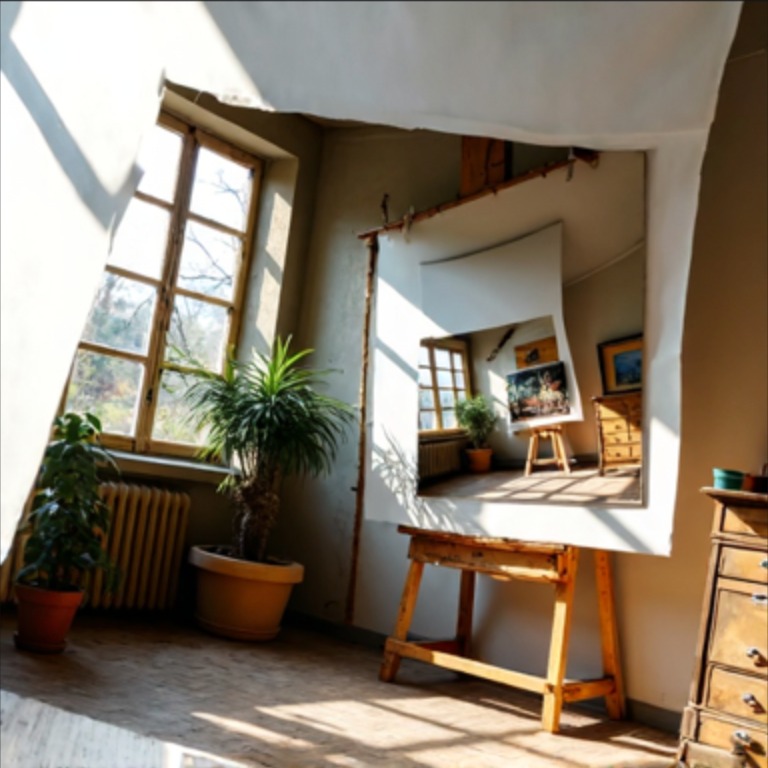}\\[\tzg]
          \includegraphics[ width=\textwidth]{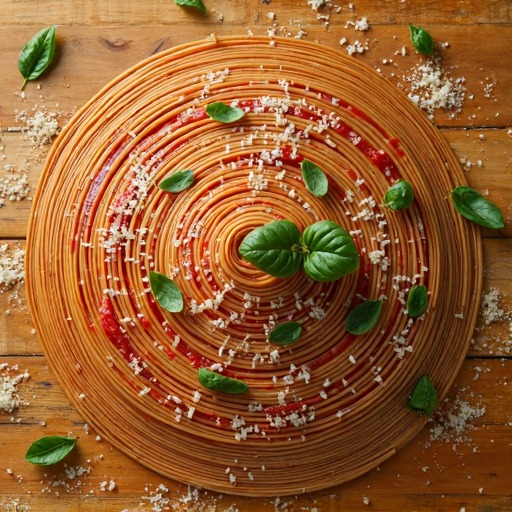}\\[\tzg]
          \includegraphics[ width=\textwidth]{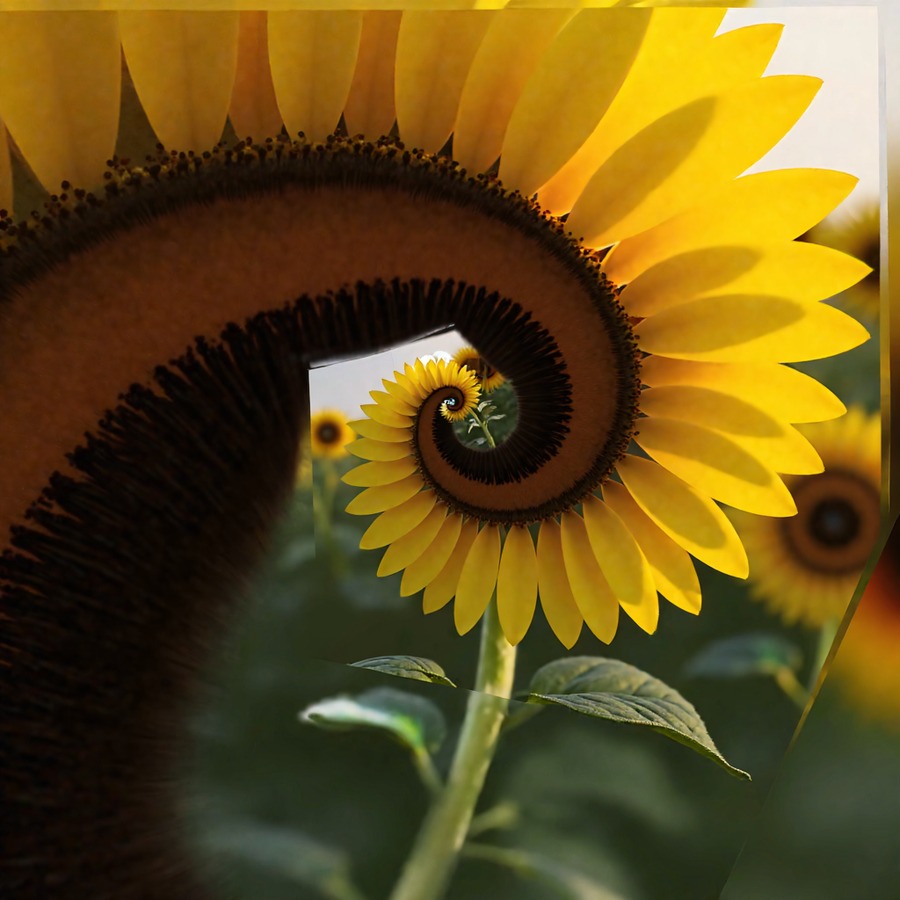}
        \end{minipage}
        \begin{minipage}[c]{\dimexpr 3\tzs+2\tzg\relax}
            \includegraphics[ width=\textwidth]{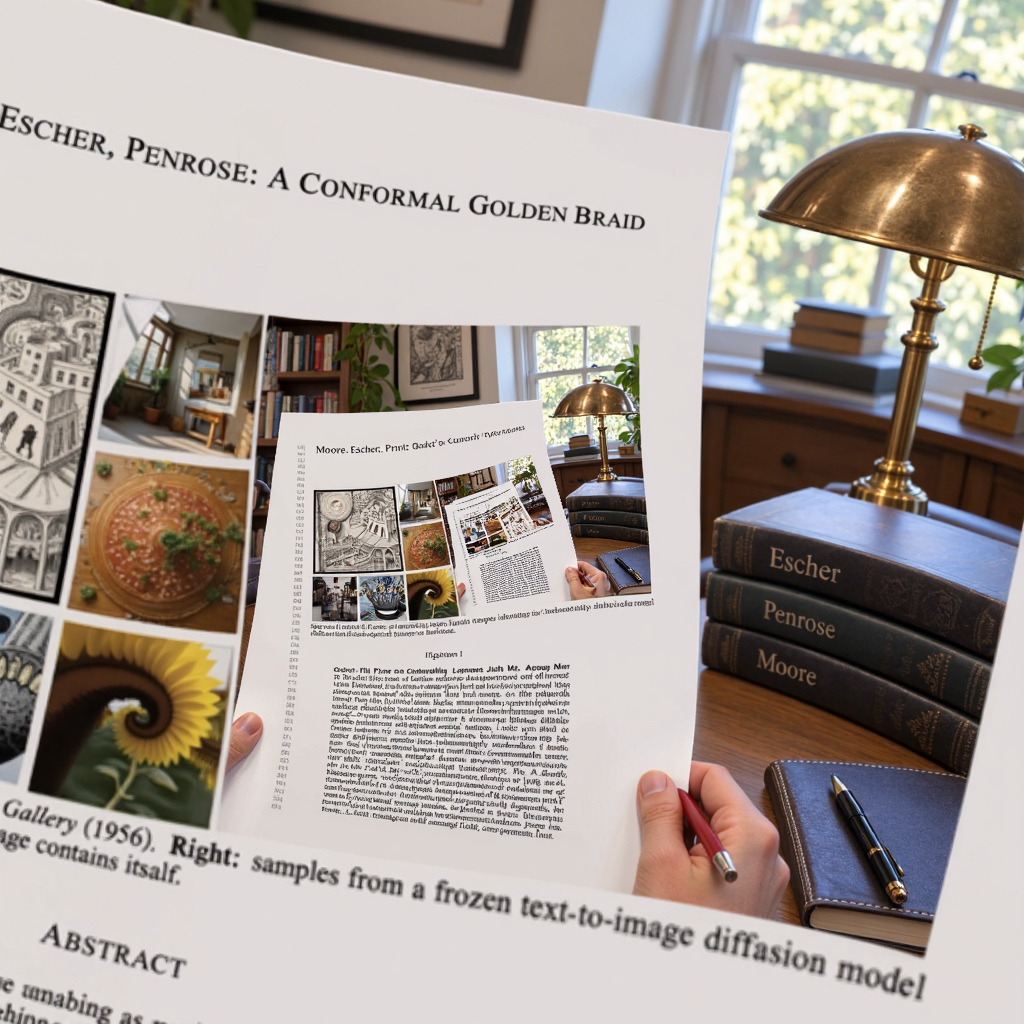}
        \end{minipage}%
      }
      
      \vspace{3pt} 
      \begin{minipage}{\dimexpr 6\tzs+5\tzg\relax}
        \captionof{figure}{\small \textbf{Top-left:} M.~C.~Escher, \emph{Print Gallery} (1956).
        \textbf{Surrounding panels:} generated recursive scenes.
        \textbf{Right:} a conditional image of a reader holding this paper,
        re-noised and refined with our braided sampler. The reader, like you,
        believes they hold the outermost copy. See the
        \textbf{animated supplementary material}.}
        \label{fig:teaser}
      \end{minipage}
    }
  }
\end{figure}

\begin{abstract}
``I don't think I have ever done anything as peculiar in my life.
Among other things, it shows a young man looking with interest at a print
on the wall of an exhibition that features himself. How can this be?
Perhaps I am not far removed from Einstein's curved universe.''
So wrote M.~C.~Escher about his 1956 lithograph \textit{Print Gallery}.
Nearly half a century later, a mathematical analysis related its geometry
to an untwisted source image through a conformal power map
$z \mapsto z^\alpha$, $\alpha \in \mathbb{C}$.
Building on this construction, we use a frozen text-to-image diffusion
model to generate new self-referential scenes.
Prompting alone does not enforce the recursion, while a post-hoc
transformation can leave structures poorly connected.
Applying the transformation during sampling is also insufficient:
the denoiser may ``repair'' the intended distortion or drift out
of the prescribed geometry.
We construct a generalized inverse $T^\dagger$ of the non-invertible
image transformation $T$, adapted to its recursive constraint.
In the idealized formulation, the Penrose identity $TT^\dagger T = T$
makes $TT^\dagger$ an idempotent projection onto geometrically
admissible images.
Yet denoising only the transformed image remains an out-of-distribution
task, even with projection.
We therefore braid denoising steps with $T$ and $T^\dagger$:
source-space steps develop the untwisted scene, while transformed-space
steps refine its appearance and connections in the final geometry.
We generate \textit{Print Gallery}-like compositions and explore
further transformations.
Rather than distorting a finished image, we let the scene and its
distortion develop together.\footnote{The title echoes Hofstadter's \emph{Gödel, Escher, Bach: an Eternal Golden Braid}~\citep{hofstadter1979}.}
\end{abstract}

\section{Introduction}
\label{sec:intro}
\begin{flushright}
    \small
    \textit{``I went there expecting staircases coming out of the windows\ldots''}\\
    Roger Penrose, recalling his 1962 visit to M.~C.~Escher
    \citep{wade2015penrose}
\end{flushright}

Maurits Cornelis Escher was not trained as a mathematician. Yet his art both
inspired mathematicians and drew on their ideas. After encountering Escher's
work, Roger Penrose began devising impossible objects. The triangle and
endless staircase he published with his father Lionel in 1958 subsequently
inspired Escher's \emph{Waterfall} and \emph{Ascending and Descending}
\citep{penrose1958,kersten2018ascending,kersten2019waterfall}.
A different loop appears in the earlier \emph{Print Gallery} (1956).
A young man studies a print of a town; following its buildings brings us
back to the gallery in which he stands. A picture contains the world that
contains it \citep{escherPrintGallery}.

\citet{desmit2003} described the print's idealized geometry through a
conformal power map connecting a scale-periodic source to its transformed
image. It specifies how coordinates correspond, not how buildings, objects, and
textures should connect. We seek to generate scenes composed within this
geometry, using a pretrained image model.

Prompting for self-reference imposes no exact relation between image regions.
Applying the transformation after generation enforces repetition, but leaves
the model no opportunity to account for the resulting connections.
Introducing it during sampling lets the model respond to the geometry,
but subsequent denoising can alter corresponding regions independently or
``repair'' the intended distortion. Here, distortion is not a defect to
remove, but part of the artistic intent.

We use two coupled representations: an untwisted, scale-recursive source and
its transformed image. A reverse coordinate warp alone is insufficient:
the image transformation selects and repeats content, and
a denoised image need not remain in its range. We therefore construct a
matched generalized inverse $T^\dagger$ that also acts on such images.
Here Penrose returns through generalized inversion
\citep{moore1920,penrose1955}. In the continuous construction,
$TT^\dagger T=T$ ensures that an admissible transformed image survives a
round trip unchanged. Thus $TT^\dagger$ is an idempotent projection onto
the range of $T$, restoring the constraint after updates,
without a Euclidean nearest-point guarantee.

Geometric consistency alone does not make a convincing composition.
Denoising only the transformed image asks the model to interpret
unfamiliar arrangements throughout generation. Braiding steps with $T$
and $T^\dagger$ lets the same frozen model develop the source scene and
its transformed appearance. For the \emph{Print Gallery} construction,
a zero-spin warm-up first encourages scale recursion.

We contribute a matched generalized inverse and a braided sampler for
recursive scene generation. The Penrose identities describe continuous
operators; resampling and subsequent denoising introduce constraint error.
Figure~\ref{fig:assorted-results} previews the range of geometries and
scenes; The \textbf{project page} provides the animated views: https://assafshocher.github.io/escher/supplement.html

\subsection{Related work}
\label{sec:related}

\begin{flushright}
    \small
    \textit{``Originality is merely an illusion.''}\\
    --- M.~C.~Escher
    \citep{escher1982life}
\end{flushright}

\paragraph{Escher's geometry and computational art.}
\citet{desmit2003} formalized the geometry of \emph{Print Gallery} and
used it to reconstruct the missing central region.
\citet{cipolinakun2022} subsequently compared CoModGANs, LaMa, and GLIDE
for completing the artwork: they straightened it into several views,
inpainted these views, and transformed the completion back.
Generation in the straightened representation is a direct precedent. We generate new scenes and repeatedly update both
representations during sampling. Generative Escher Meshes \mbox{\citep{escherMeshes2024}} combines diffusion guidance
with geometric constraints to synthesize tileable shapes and textures.

\paragraph{Self-similarity and generation across scales.}
Iterated function systems and fractal image coding represent structure
through repeated contractive maps \citep{hutchinson1981,jacquin1992}.
Neural Collages \citep{poli2022} learns contractive operators whose fixed
points represent images. Our geometry prescribes admissible images, with content from a frozen model.
Generative Powers of Ten \citep{powersoften2024} coordinates diffusion
across scales for semantic zooms; ZoomLDM \citep{zoomldm2025}
learns generation at multiple magnifications. These methods allow content
to change with scale, whereas our construction repeats the same content. Fractal Generative Models \citep{li2025fractal} uses
architectural recursion without imposing this equality on outputs.

\paragraph{Constrained diffusion sampling.}
DDRM \citep{kawar2022} exploits the spectral structure of linear inverse
problems, while DDNM \citep{wang2023ddnm} corrects clean-image estimates
using a generalized inverse. DPS \citep{chung2023} uses approximate
likelihood guidance for noisy linear and nonlinear measurements.
$\Pi$GDM \citep{song2023pigdm} also uses pseudoinverse-based guidance,
including nonlinear and nondifferentiable measurement operators.
Universal Guidance, Self-Guidance, and fast constrained sampling provide
further ways to steer a frozen prior
\citep{bansal2023,epstein2023,fast2025}.
Here the constraint relates regions of the generated image
to one another, and sampling alternates between source and transformed
representations using the matched operators.

\paragraph{Coupled views and geometric consistency.}
MultiDiffusion \citep{bartal2023} fuses diffusion paths under shared
constraints, and SyncTweedies \citep{synctweedies2024} synchronizes
clean-image estimates through a canonical representation.
Visual Anagrams \citep{geng2024anagrams} combines denoising predictions
under orthogonal image transformations. LookingGlass
\citep{lookingglass2025} extends this approach to latent flow models and
more general spatial warps through Laplacian-pyramid image processing.
Diffusion Illusions \citep{burgert2024} optimizes images whose arrangements
reveal different interpretations. Tiled Diffusion \citep{tiled2025}
enforces compatibility between tile boundaries, including self-tiling;
Refracting Reality \citep{refracting2026} synchronizes directly and
indirectly visible content using optical warps. These methods establish denoising across coupled views; our contribution
is a matched operator pair for scale-recursive geometry and its use in
sequential source/Escher sampling. The forward operator discards content
outside a selected source band, while the generalized inverse selects
its Droste extension. Their composition is an idempotent projector onto
the twisted image class, allowing the constraint to be restored without
compounding the warp. Braided sampling inserts source-space denoising
between the inverse and forward maps to develop the scene within this
recursive geometry.

\paragraph{Idempotence and generalized inversion.}
Idempotent Generative Network \citep{shocher2024ign} trains a generator
using an idempotence objective; here the geometric operator is specified
analytically and the image prior is frozen.
\citet{gofer2024} develop generalized inverses for nonlinear operators,
including a norm-based pseudoinverse, while \citet{ehrlich2026spnn}
construct surjective pseudo-invertible neural networks (SPNNs) and define
inversion through a bijective completion. These formulations distinguish algebraic consistency from
the criterion selecting an inverse.

\section{Preliminaries}
\label{sec:prelim}

\subsection{From scale repetition to Escher's geometry}
\label{sec:desmit}

The \emph{Droste effect} is the repeated appearance of an image
within itself.\footnote{The name comes from Droste cocoa
packaging, which depicts a nurse carrying a tray with a smaller
package bearing the same image.}
Following \citet{desmit2003}, we start with repetition under scaling alone, then apply
a conformal power map to introduce rotation.
Place the recursion centre at the origin and represent image
locations by complex numbers $z=x+iy=re^{i\theta}$.
Choose a scale factor $\lambda>1$ and write $L=\log\lambda$.
Our source images satisfy $s(\lambda z)=s(z)$:
zooming by $\lambda$ reproduces the same image.
Denote this source class by $\mathcal S$.

\paragraph{Power map.}
We transform source $z$ into output $w$
using a complex power on a logarithm branch:
\[
  w=z^\alpha=\exp(\alpha\log z).
\]
Locally, away from the origin, this map is \textbf{conformal}: it preserves
angles.
Write $\log z=\log r+i\theta$.
In these coordinates, scaling becomes a horizontal shift and
rotation becomes a vertical shift. Multiplication by $\alpha$ rotates and scales this plane;
exponentiation returns us to image coordinates.

\paragraph{Choosing the exponent.}
An output point must receive the same colour at angles $\theta$ and $\theta+2\pi$.
For the \emph{Print Gallery} construction, one full turn in the
output corresponds to one full turn plus one scale step in the
source. Both leave the source image unchanged.
Since the inverse coordinate map multiplies logarithmic
coordinates by $\beta=\alpha^{-1}$, this requires
$\beta(2\pi i)=2\pi i+L$, giving
\begin{equation}
  \beta=1-i\frac{L}{2\pi},
  \qquad \alpha=\beta^{-1}.
  \label{eq:powermap}
\end{equation}

\begin{figure}[t]
  \centering
  \includegraphics[ width=0.8\linewidth]{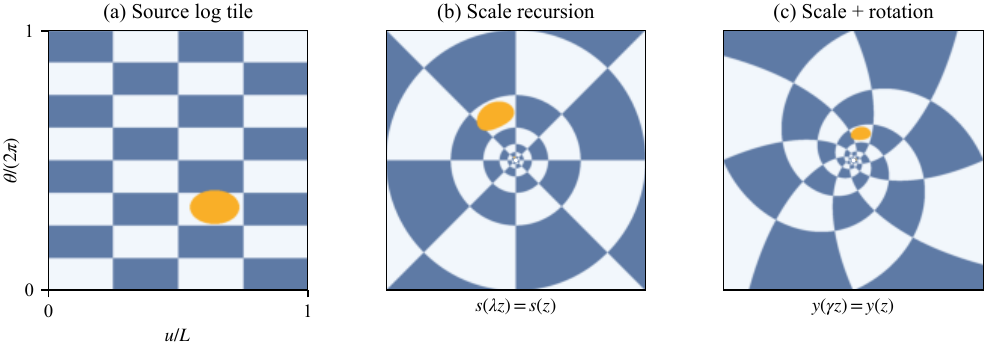}
\vspace*{-0.15cm}\caption{\small \textbf{The Droste effect in three representations.}
Left: a tile repeated in logarithmic coordinates, with horizontal
period $L=\log\lambda$ and vertical period $2\pi$.
Middle: returning to image coordinates gives scale repetition,
$s(\lambda z)=s(z)$.
Right: the conformal power map adds rotation, producing the
Escher geometry with $y(e^{\alpha L}w)=y(w)$.
All three panels use the same pattern, with $\lambda=16$.}
\label{fig:prelim}
\end{figure}

\paragraph{From coordinates to images.}
A source band $R/\lambda<|z|\le R$ determines the scale-recursive
image by repetition inward and outward. Each output location reads
its colour from the corresponding source location:
\begin{equation}
  y(w)=s\!\left(\exp(\beta\log w)\right),
  \qquad
  y(e^{\alpha L}w)=y(w).
  \label{eq:recursive-image}
\end{equation}
The choice of $\beta$ makes this sampling rule independent of
the logarithm branch. Multiplication by $e^{\alpha L}$ advances
one source scale period, reproducing the same image.
In the output, this corresponds to scaling by
$e^{L\operatorname{Re}\alpha}$ and rotating by
$L\operatorname{Im}\alpha$ radians.
These images form the transformed class $\mathcal E$.
For $\lambda=256$, the smaller copy has scale approximately
$1/22.58$ and rotates clockwise by $157.63^\circ$
\citep{desmit2003}.
The construction excludes the origin; finite images sample
a bounded window around it. Figure~\ref{fig:prelim} compares these three representations.

\subsection{Generalized inverses and idempotent projections}
\label{sec:generalized-inverses}

For $A\in\mathbb C^{m\times n}$, the Moore--Penrose
inverse $A^\dagger$ uniquely satisfies
\citep{moore1920,penrose1955}:
\begin{equation}
\begin{aligned}
AA^\dagger A&=A,
& A^\dagger AA^\dagger&=A^\dagger,\\
(AA^\dagger)^*&=AA^\dagger,
& (A^\dagger A)^*&=A^\dagger A,
\end{aligned}
\label{eq:four-penrose}
\end{equation}
where $*$ denotes the conjugate transpose.
It returns the minimum-norm least-squares solution.

Keeping only the first two identities defines a
\emph{reflexive generalized inverse} $A^\ddagger$, which
need not be unique. The round trips are already idempotent:
$
  AA^\ddagger AA^\ddagger=AA^\ddagger,
  \quad
  A^\ddagger AA^\ddagger A=A^\ddagger A.
  $
The last two Penrose identities additionally make these
projections orthogonal.

\paragraph{Nonlinear inverses.}
For nonlinear maps, the first two identities still make sense:
$
  gg^\dagger g=g,
  \qquad
  g^\dagger gg^\dagger=g^\dagger.
$
The last two rely on a linear adjoint and require a replacement.
\citet{gofer2024} retain reflexive consistency and select
$g^\dagger(y)$ by first minimizing $\|g(x)-y\|$, then
minimizing $\|x\|$ among those minimizers.
This recovers the Moore--Penrose inverse in Euclidean linear
spaces.

For surjective $g$, \citet{ehrlich2026spnn} fix a
bijective completion $G(x)=(g(x),q(x))$ and select
\begin{equation}
  g^\dagger(y)=
  \operatorname*{arg\,min}_{x:\,g(x)=y}
  \|G(x)-G(0)\|_2.
\end{equation}
Thus distance is measured in the completed coordinates.
For $g=A\circ\phi$, with $\phi$ bijective and $\phi(0)=0$,
a completion matched to these linearizing coordinates gives
$g^\dagger=\phi^{-1}\circ A^\dagger$, recovering all four
Penrose identities in those coordinates.
Their SPNNs implement this construction.

\paragraph{Our inverse convention.}
We use $T^\dagger$ to denote a chosen reflexive generalized
inverse, selected by the scale-repeating Droste structure
and matched fundamental bands.
The dagger does not assert the Moore--Penrose minimum-norm
or orthogonality properties.
Its round trips restore recursive image structure; here, \emph{projection} means idempotence,
without a nearest-point claim.

\section{Method}
\label{sec:method}
\begin{figure}[t]
  \centering
  \includegraphics[width=\linewidth]{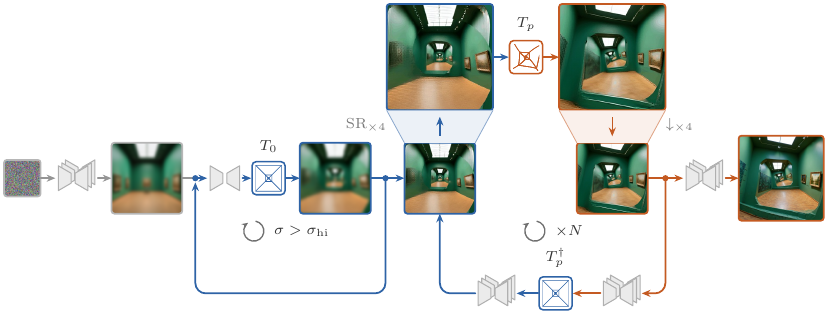}
  \vspace*{-0.15cm}\caption{\small \textbf{Braided sampling.}
  Warm-up interleaves denoising with untwisted repetition $T_0$.
  The sampler switches between Droste and Escher representations
  using $T$ and $T^\dagger$, with denoising blocks $D^{(n)}$ between
  switches. It ends with refinement in the Escher representation. At each switch, $4\times$ super-resolution precedes the map,
  then resizing to working resolution.
  Dashed arrows indicate repetition; the noise-level axis is
  schematic. The early Droste preview is an illustrative blur of the
  later panel, not a recorded checkpoint.}
  \label{fig:method}
\end{figure}

We generate recursive scenes with a frozen text-to-image model by
applying geometric maps to its predicted clean images between sampling
steps. Alternating denoising with $T$ and its matched generalized inverse
$T^\dagger$ develops the untwisted scene and the connections created by
its Escher transformation. Figure~\ref{fig:method} shows this
\emph{braided sampling} process, from scale-repetition warm-up to
final refinement in the transformed representation.

\subsection{Moving between representations}
\label{sec:pair}

The maps must accept images that violate recursion, since
denoising can change corresponding copies independently.
Let $B_1$ extract the chosen source band and $E_1$ repeat its content
under scaling. Likewise, $B_2$ reads the corresponding band in the
Escher image, and $E_2$ renders it with the scale and rotation from
Sec.~\ref{sec:desmit}. Both pairs use the same band coordinates.
In the continuous construction, extracting a rendered band returns
its content unchanged:
$B_1E_1=B_2E_2=\mathrm{id}$. Define
\begin{equation}
  T=E_2B_1,
  \qquad
  T^\dagger=E_1B_2.
  \label{eq:pair}
\end{equation}
Each map keeps the selected band and regenerates the remaining image.
This explains both its loss of information on arbitrary inputs and
its ability to impose recursion in the destination representation.

\paragraph{Why the inverse is matched.}
Cancelling $B_1E_1$ and $B_2E_2$ in these compositions gives
\begin{equation}
\begin{aligned}
  TT^\dagger T&=T,
  &T^\dagger TT^\dagger&=T^\dagger,\\
  TT^\dagger&=E_2B_2,
  &T^\dagger T&=E_1B_1.
\end{aligned}
  \label{eq:method-round-trips}
\end{equation}
Thus an admissible image survives a geometric round trip unchanged.
For an arbitrary image, the round trip restores repetition from its
selected band. On the admissible classes $\mathcal S$ and $\mathcal E$,
the maps are mutual inverses; on the ambient image spaces they are
reflexive generalized inverses. The band boundaries and angular origin
must agree between extraction and rendering.

\paragraph{Restoring the constraint without compounding the warp.}
Applying $T$ again to an Escher image generally compounds the coordinate
warp; $T$ itself is not the projector onto the Escher class.
The matched pair instead gives $P_{\mathcal E}=TT^\dagger=E_2B_2$,
which is idempotent and fixes every admissible Escher image in the
continuous construction. The inverse selects a Droste representative:
it reads the chosen Escher band, untwists it, and extends it by scale
repetition. Appendix~\ref{app:droste-inverse} gives an explicit
construction. An immediate $T^\dagger$--$T$ round trip restores the
constraint; our sampler additionally denoises in the Droste
representation between these maps to update the scene content.

\paragraph{Two paths for updating the content.}
The model can update a Droste image directly, or update its Escher
rendering and carry the changes back:
\begin{equation}
  s \xrightarrow{T} y
    \xrightarrow{D_{\mathcal E}} \widetilde y
    \xrightarrow{T^\dagger} s'.
  \label{eq:dual-path}
\end{equation}
Here $D_{\mathcal E}$ denotes denoising in the Escher representation.
The two paths develop the untwisted scene and its warped connections.
Each geometric map ideally restores recursion in its destination,
without an additional round-trip projection at the switch.

\subsection{Sampling schedule}
\label{sec:sampler}

We use the same frozen model and prompt in both representations.
In the schematic sequence
\begin{equation}
  \cdots \to D_{\mathcal S}\to T
  \to D_{\mathcal E}\to T^\dagger
  \to D_{\mathcal S}\to T\to\cdots,
  \label{eq:opt1}
\end{equation}
each $D$ denotes a block of sampling steps. Geometric operations act
on the model's predicted clean image, rather than on its noisy input.

\paragraph{Starting with scale repetition.}
At high noise levels, we apply the untwisted map
$T_0=E_1B_1$ after each clean prediction, at the working resolution.
This introduces nested copies as the scene develops.
Despite its identity coordinate warp, $T_0$ selects and repeats a band. Warm-up ends at $\sigma_{\mathrm{hi}}$.

\paragraph{Switching representations.}
The default conformal schedule applies $T$ at $\sigma_{\mathrm{hi}}=0.87$,
$T^\dagger$ nine steps later, and a final $T$ at
$\sigma_{\mathrm{lo}}=0.5$, then finishes denoising in the Escher
representation. The gap $g$ controls only the first two switches;
Appendix~\ref{app:sampler} gives the full schedule.

\paragraph{Updating the noisy state.}
Only when a geometric map is applied, including warm-up, do we
re-noise the modified clean prediction with fresh noise:
\begin{equation}
  h_{\mathrm{next}}
  =a_{\mathrm{next}}\widetilde h_{\mathrm{clean}}
   +b_{\mathrm{next}}\epsilon,
  \qquad \epsilon\sim\mathcal N(0,I).
  \label{eq:method-renoise}
\end{equation}

Here $h$ is the sampling state and $a,b$ are the schedule's signal
and noise coefficients. Other steps use the regular sampler update.
Fresh noise avoids warping existing noise: a linear resampler $W$
would change its covariance from $I$ to $WW^*$.

\subsection{Resolution and transformation choices}
\label{sec:latent}

\paragraph{Resolving quality and resolution.}
At each switch, we apply the geometric map to a $4\times$
super-resolved pixel image, then resize it to the working resolution:
\begin{equation}
  M_{\mathrm{pix}}(x)=
  \operatorname{down}_4\!\left(M(\operatorname{SR}_4(x))\right),
  \qquad M\in\{T,T^\dagger\}.
  \label{eq:method-sr}
\end{equation}
Super-resolution provides more samples for compressed inner copies.
For latent backbones, each switch decodes, super-resolves, maps,
resizes, and re-encodes the prediction.
As an optional final step we re-noise the finished estimate to a
moderate level ($\sigma_{\mathrm{tt}}=0.75$) and denoise it once more
without operators, so the prior can smooth the residual seams the warps may
leave between repeated copies.

\paragraph{Varying the geometry.}
We explore five transformation families that arrange the
scale-repeating source content in different geometries.
For the conformal twist, $\beta_p=1-ipL/(2\pi)$ advances
$p$ source scale periods per output revolution; changing
the sign of $p$ reverses the winding direction.
\emph{Poles} combines twists around several prescribed centres,
while \emph{M\"obius} applies a M\"obius coordinate change
before the twist, producing two linked recursion centres.
\emph{Square} uses square-perimeter coordinates to produce
square-shaped nesting, and \emph{Rimrings} adds a radial term
that packs repetitions near an outer circle.
For Rimrings, we omit inverse steps because inversion is
numerically unstable; visible divisions between repeated bands
are intentional in this setting.
Appendix~\ref{app:transformations} gives the map definitions
and rendering details.

\section{Experiments}
\label{sec:results}
\label{sec:experiments}

\begin{figure}[!t]
  \centering
  \captionsetup{skip=3pt}
  \includegraphics[width=1.1\linewidth]{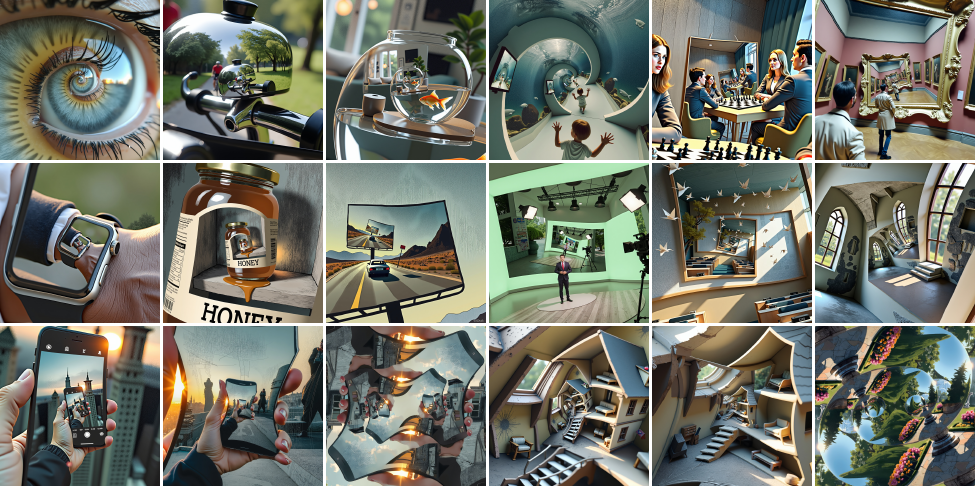}
  \caption{\small\textbf{Overview of generated results.}
    Diverse scenes with conformal twists, square spirals, and multi-center
    maps. See the \textbf{animated supplementary material}.}
  \label{fig:assorted-results}
\end{figure}

Figure~\ref{fig:assorted-results} opens with results across scenes and
transformation families. We evaluate qualitatively, inspecting recursive
layout and the content within smaller copies. All experiments use a
frozen text-to-image backbone. Baselines and ablations hold the base
prompt and initial seed fixed within each row, apart from the stated
prompt extension. Appendices~\ref{app:impl} and~\ref{app:idempotence}
give comparison conventions and a geometric round-trip check.
The \textbf{supplementary material} animates repeated views of the
images without generating new content. See
Appendix~\ref{app:additional-results} for further results.

\begin{figure}[h!]
\vspace{-0.3cm}
  \centering
  \includegraphics[ width=1.1\textwidth]{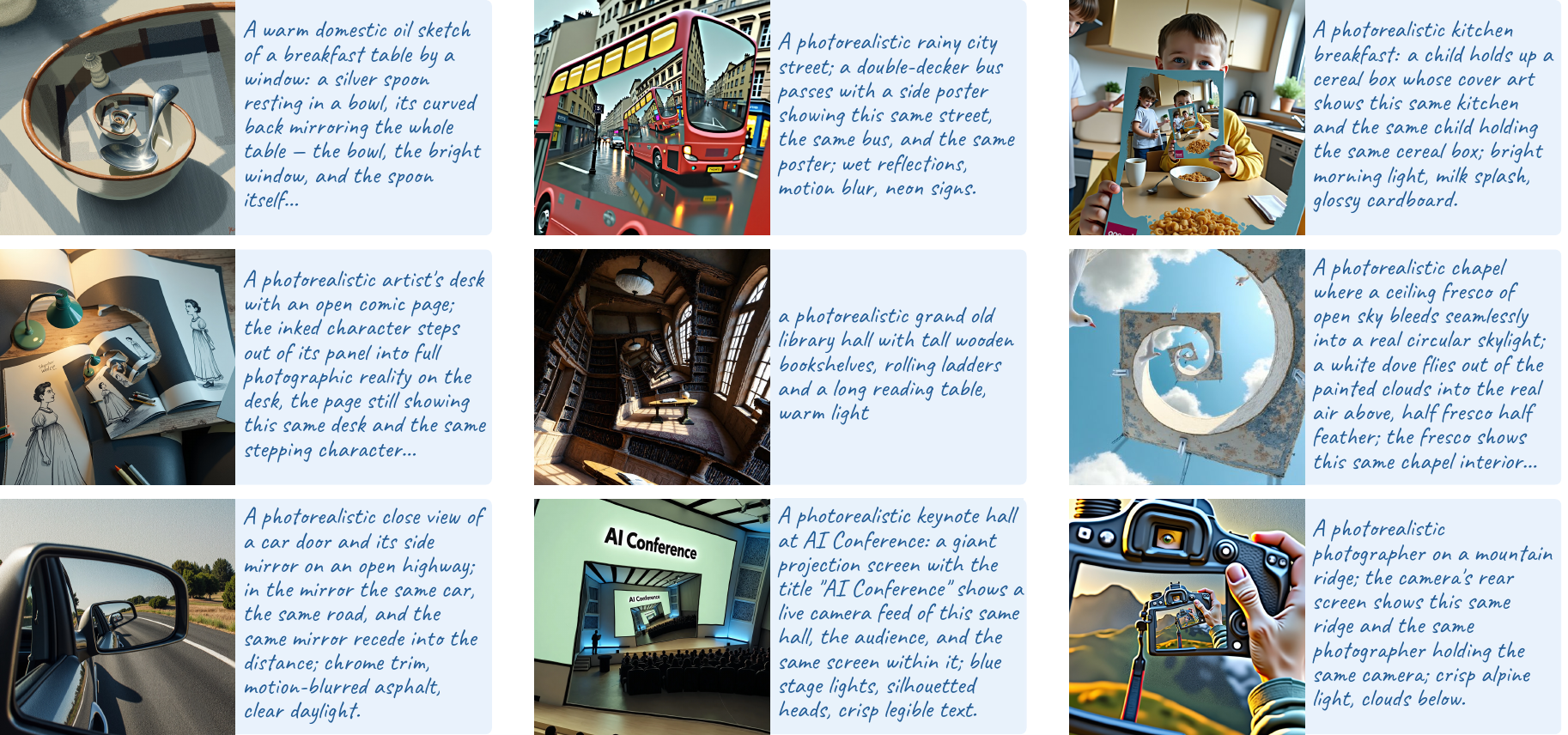}
  \vspace*{-0.15cm}\caption{\small\textbf{Conformal results with input prompts.}
  Longer prompts are excerpted. Animated views appear in the
  \textbf{supplementary material}.}
  \label{fig:general-results}
\end{figure}

\vspace{-0.25cm}\subsection{Classical Print Gallery transformation}

Figure~\ref{fig:general-results} focuses on the conformal construction,
with prompts or excerpts. Recursion follows architectural frames,
mirrors, camera displays, and screens. The library prompt requests an
ordinary reading room; geometry supplies the nesting. See the
\textbf{animated supplementary material} and
Appendix~\ref{app:additional-results} for further views and examples.

\subsection{Twist and transformation choices}

\begin{figure}[t]
  \centering
  \includegraphics[ width=1.1\linewidth]{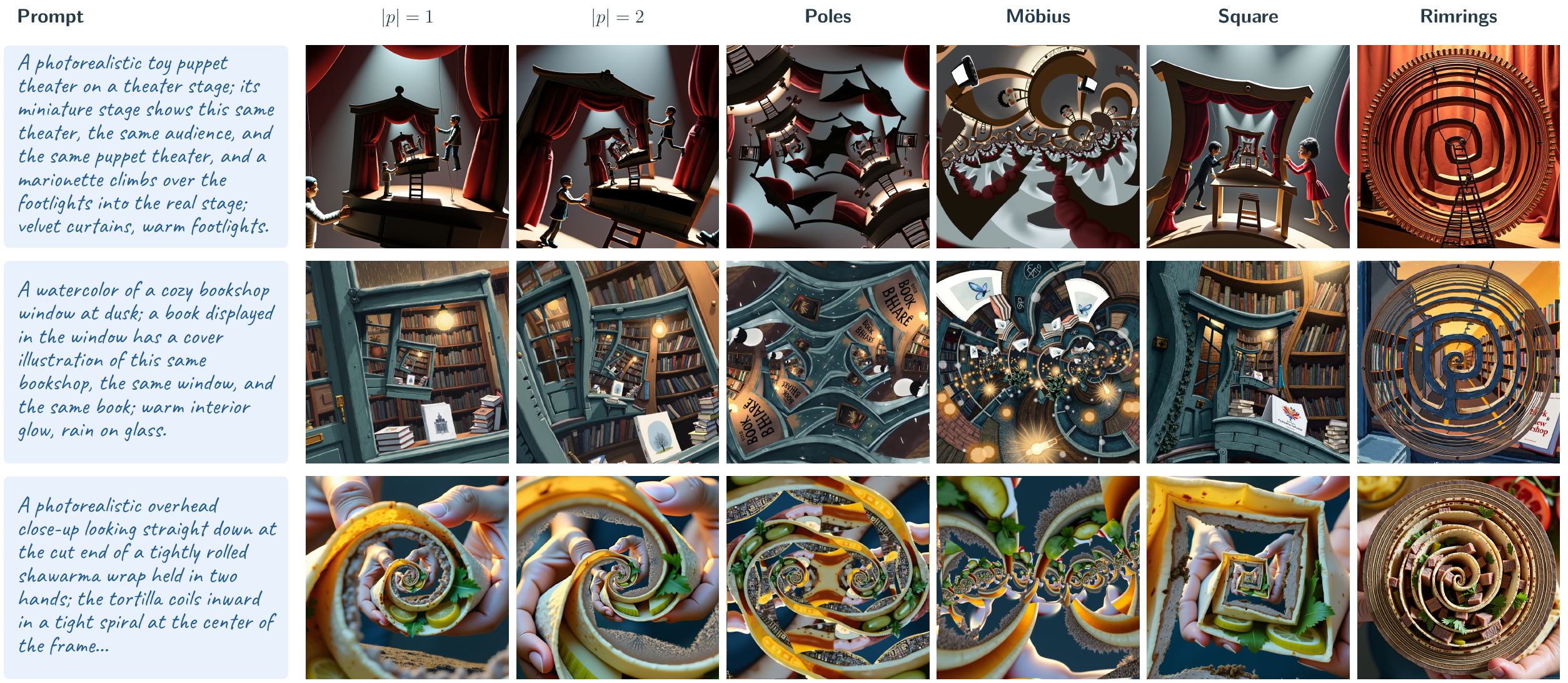}
  \vspace*{-0.15cm}\caption{\small\textbf{Twist and transformation choices.}
  Each row shares a prompt and initial seed. The conformal columns use
  $|p|=1,2$ (negative $p$ for the bookshop), followed by Poles,
  M\"obius, Square, and Rimrings. Longer prompts are excerpted.
  Rimrings omits inverse steps. See the \textbf{supplementary material}
  for animated views.}
  \label{fig:transformations}
  \label{fig:hyperparameters}
\end{figure}

Figure~\ref{fig:transformations} varies the conformal twist and
transformation family for three fixed prompt--seed pairs. Poles and
M\"obius repeat motifs around two centres, while Square produces
roughly rectangular nesting. Across these maps, recognizable scene
elements persist as their arrangement changes.

\begin{figure}[t]
  \centering
  \includegraphics[ width=0.15\linewidth]{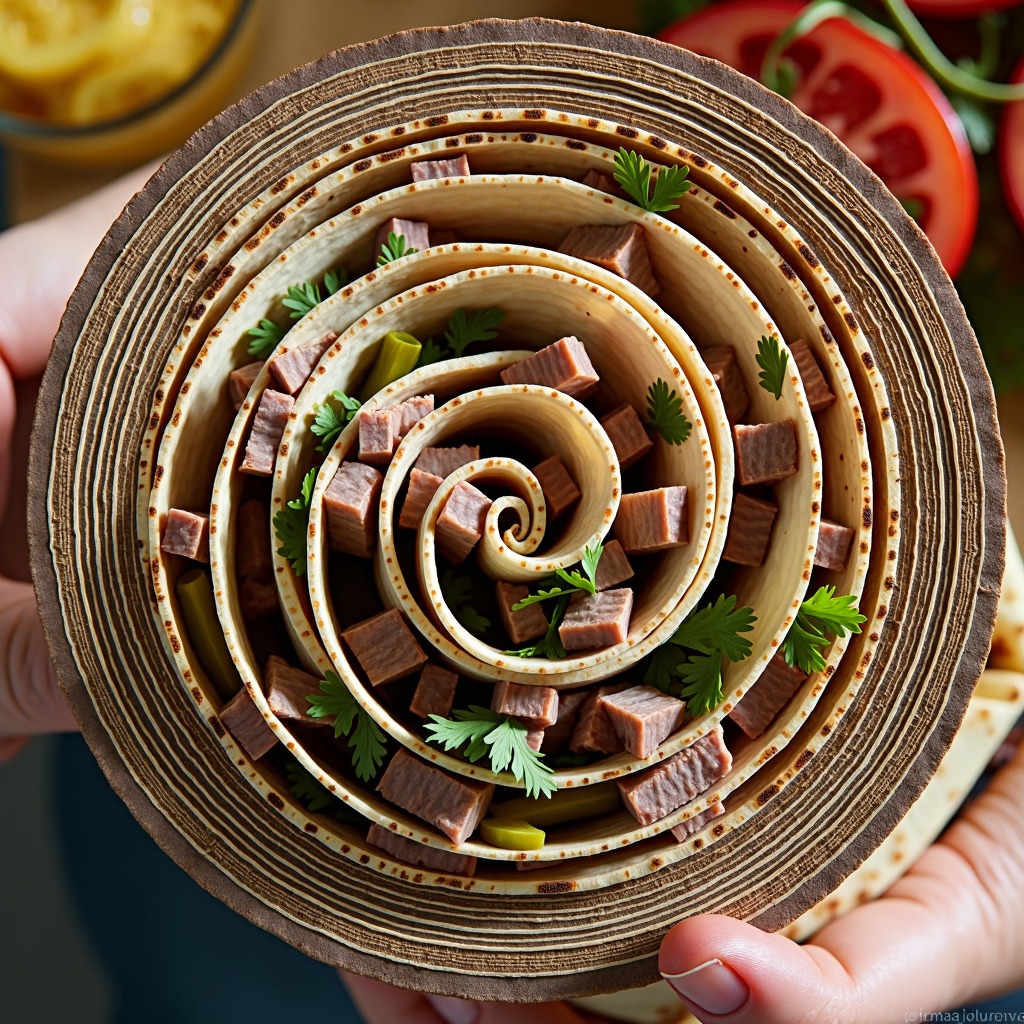}\hfill%
  \includegraphics[ width=0.15\linewidth]{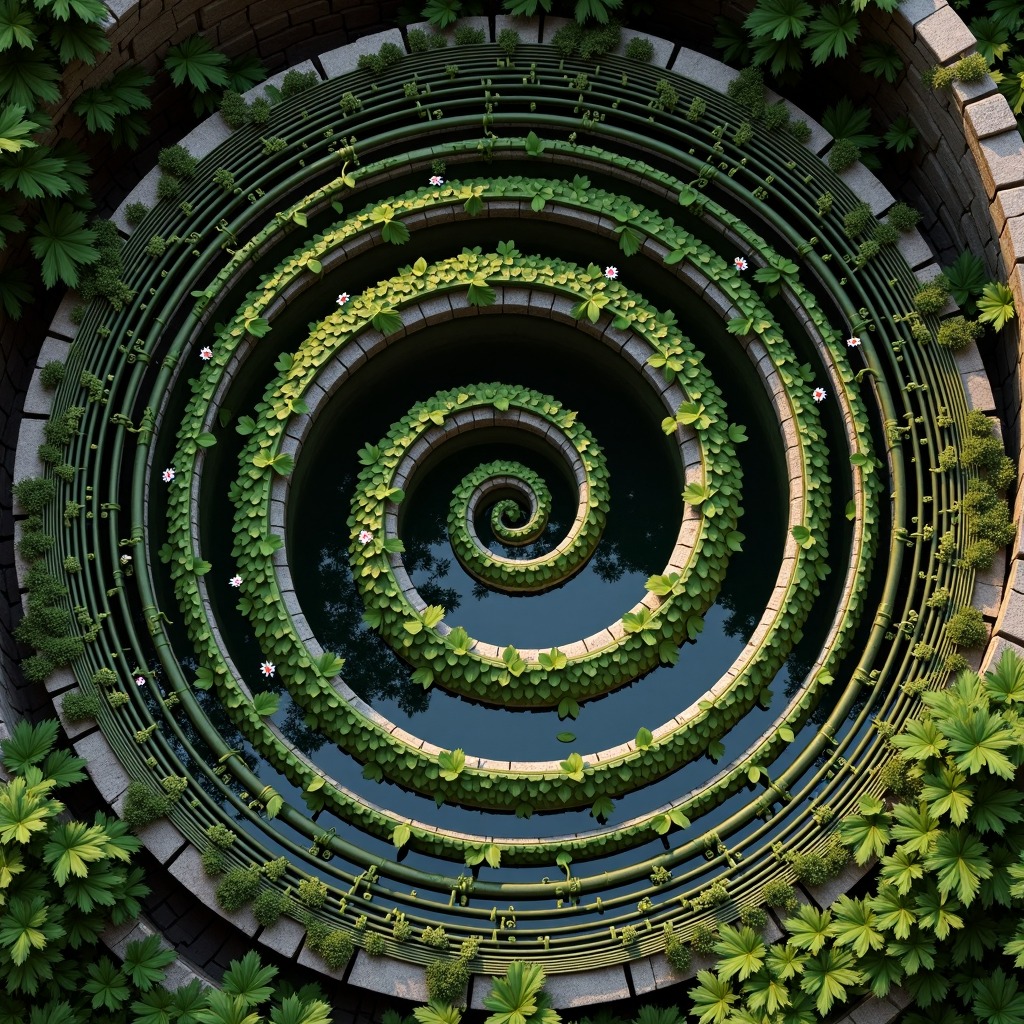}\hfill%
  \includegraphics[ width=0.15\linewidth]{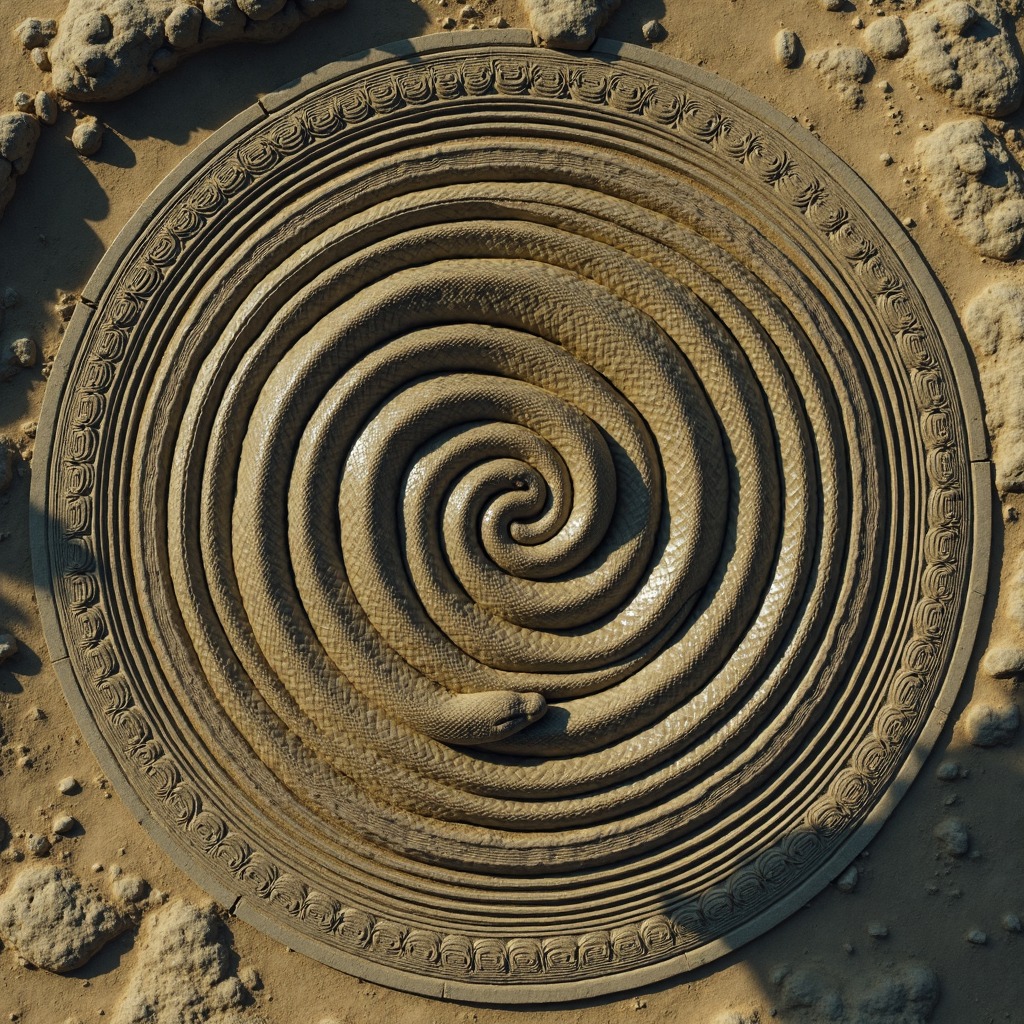}\hfill%
  \includegraphics[ width=0.15\linewidth]{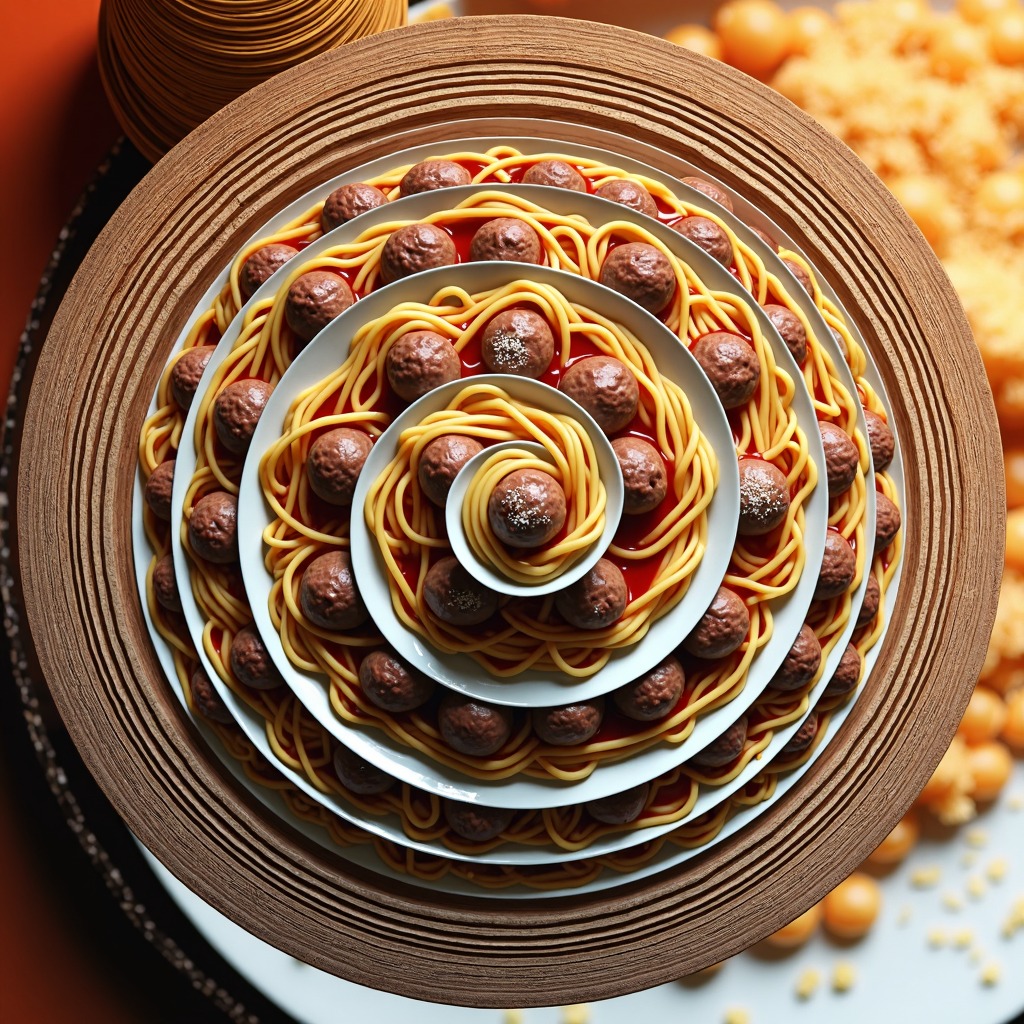}\hfill%
  \includegraphics[ width=0.15\linewidth]{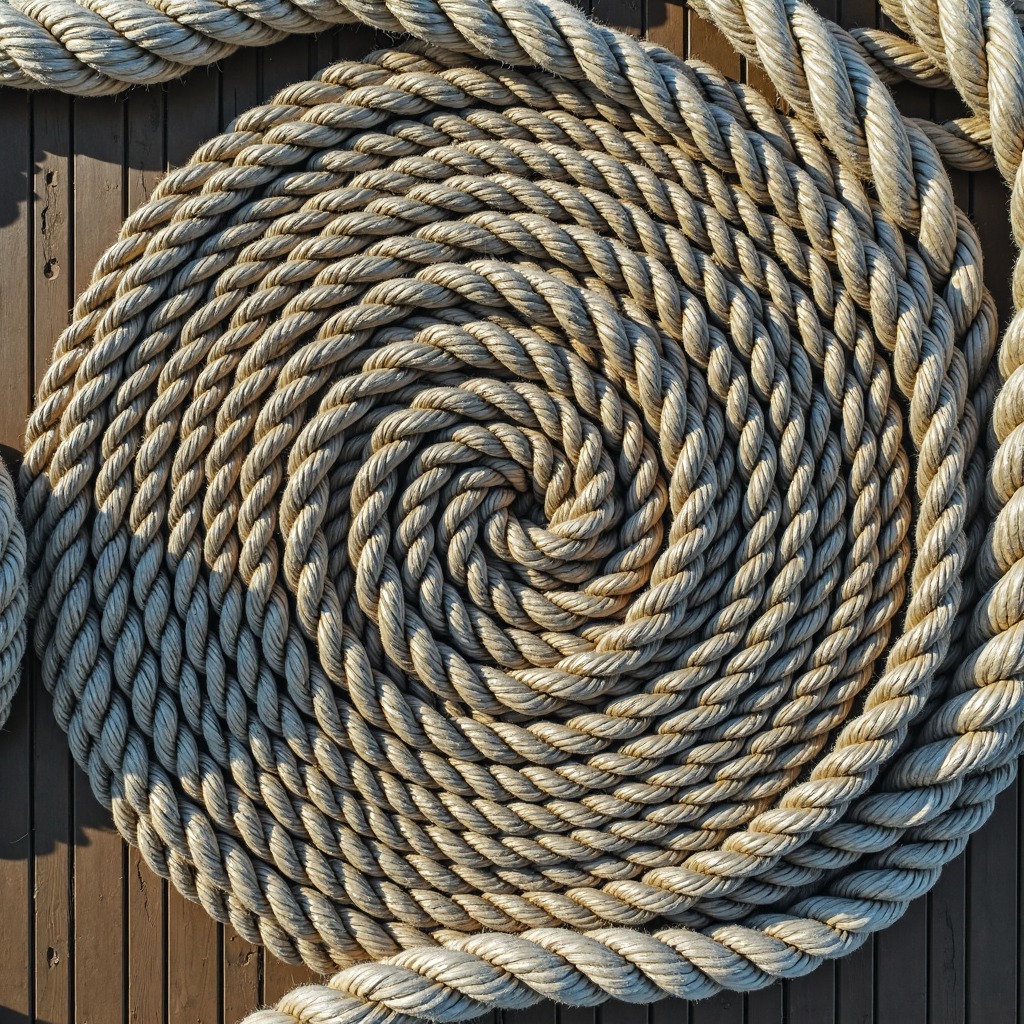}\hfill%
  \includegraphics[ width=0.15\linewidth]{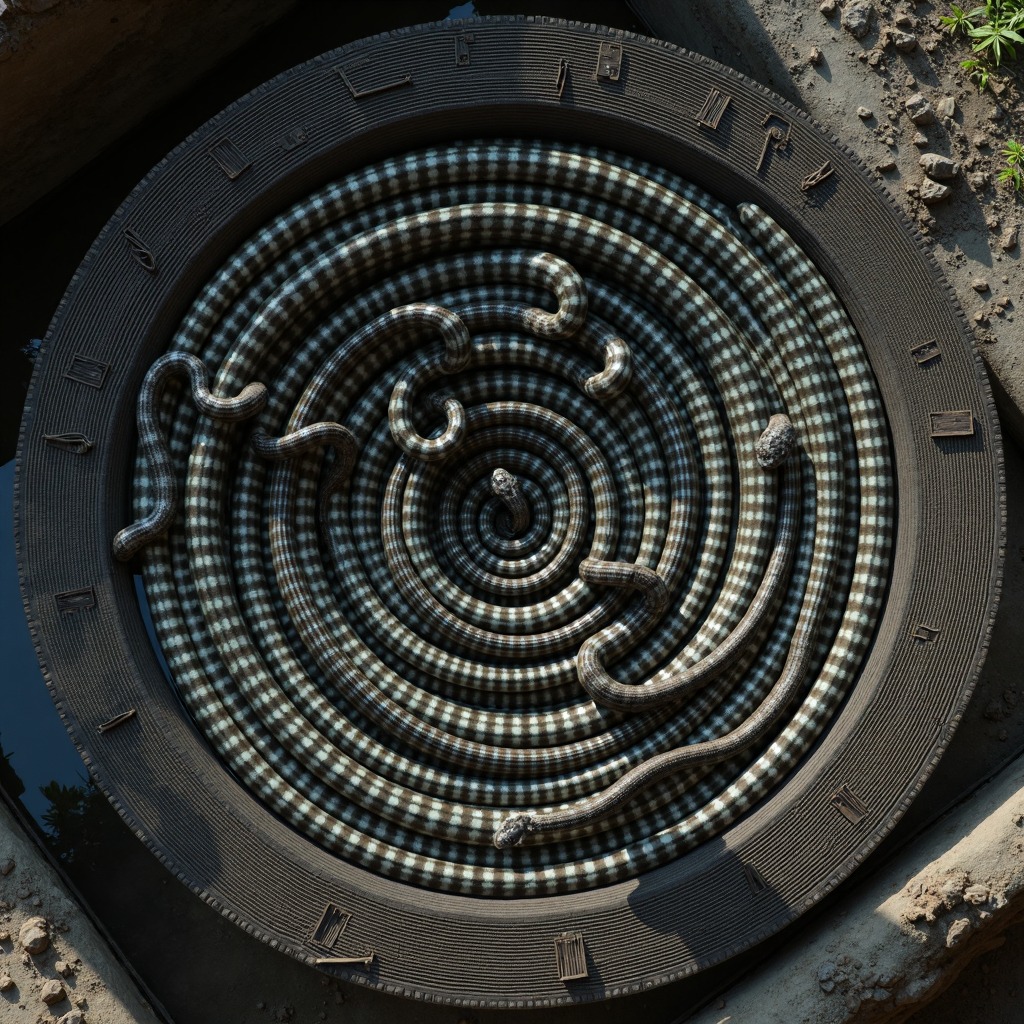}
  \caption{\small\textbf{Rimrings: repetition with visible boundaries.}
  Bands accumulate near an outer rim; their divisions form material
  layers, ledges, coils, or plate edges. These examples omit the unstable
  inverse. Animated views appear in the \textbf{supplementary material}.}
  \label{fig:rimrings}
\end{figure}

Figure~\ref{fig:rimrings} shows bands forming tortilla layers, garden
ledges, coils, and plate edges. Rimrings omits the unstable inverse:
these divisions are intentional, so the aim is not to merge them into
a continuous scene. The \textbf{supplementary material} animates this
accumulation toward the rim.

\subsection{Baselines and component ablations}
\label{sec:ablation}

\begin{figure}[t]
  \centering
  \includegraphics[ width=1.1\linewidth]{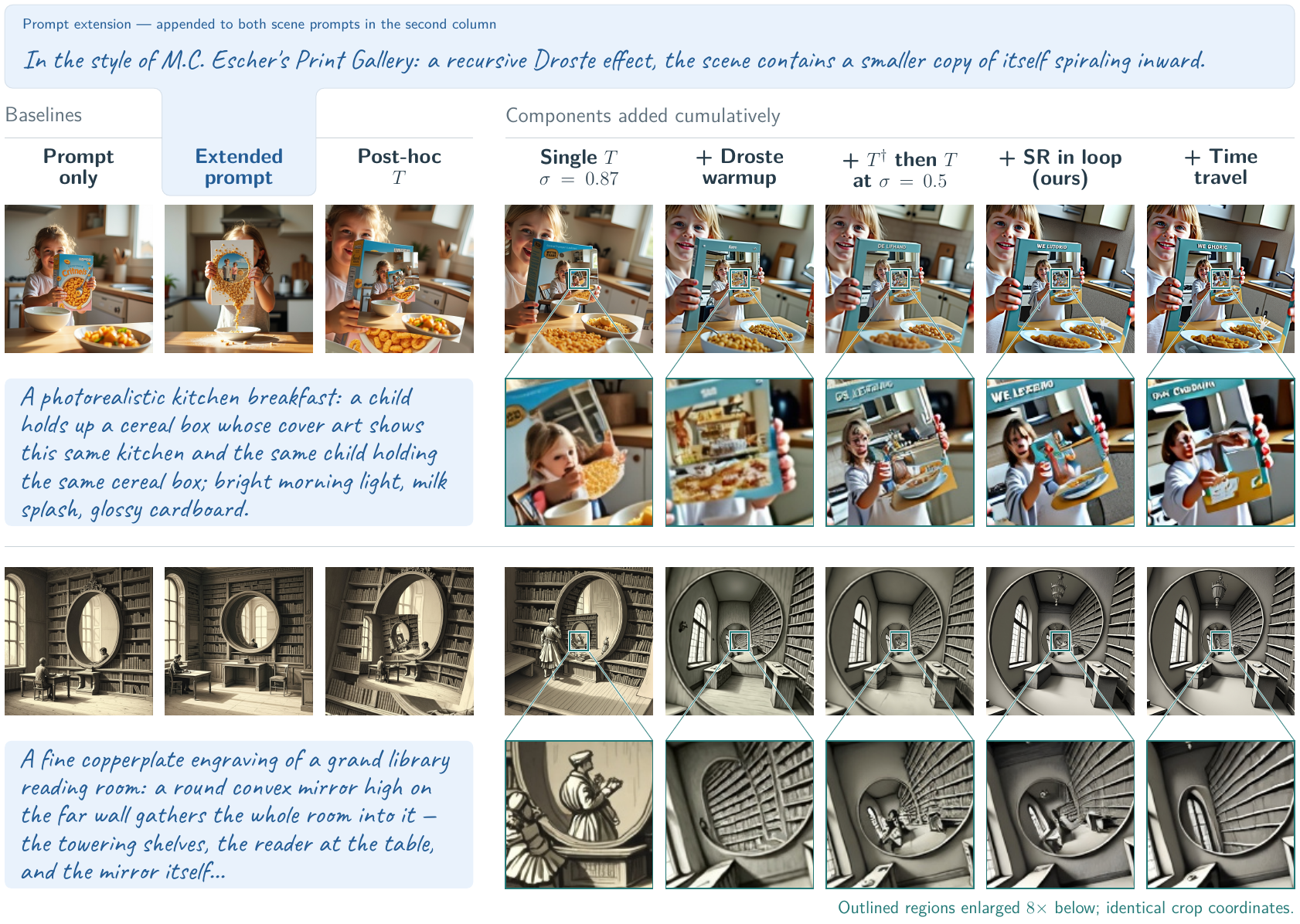}
  \vspace*{-0.15cm}\caption{\small\textbf{Baselines and cumulative ablation.}
  Rows fix the base prompt and seed; the blue callout supplies the prompt
  extension. The first three columns show prompt-only, extended-prompt,
  and post-hoc $T$ baselines. Starting from $T$ at $\sigma=0.87$,
  subsequent columns add warm-up, $T^\dagger$ followed by a final $T$
  at $\sigma=0.5$, in-loop SR (ours), and optional time travel.
  Identical central crops are enlarged $8\times$ below.}
  \label{fig:ablation}
\end{figure}

\begin{figure}
  \centering
  \includegraphics[ width=0.14\linewidth]{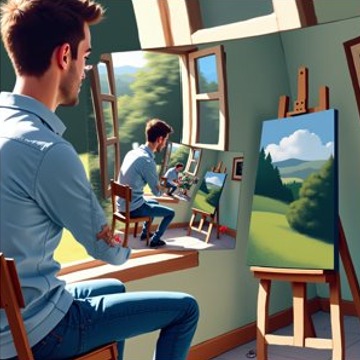}%
  \hspace{0.02\linewidth}%
  \includegraphics[ width=0.14\linewidth]{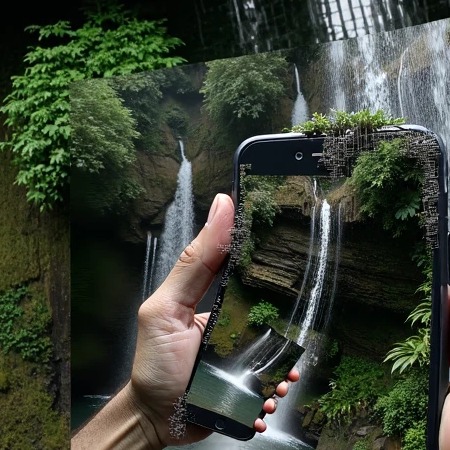}%
  \hspace{0.005\linewidth}%
  \includegraphics[ width=0.14\linewidth]{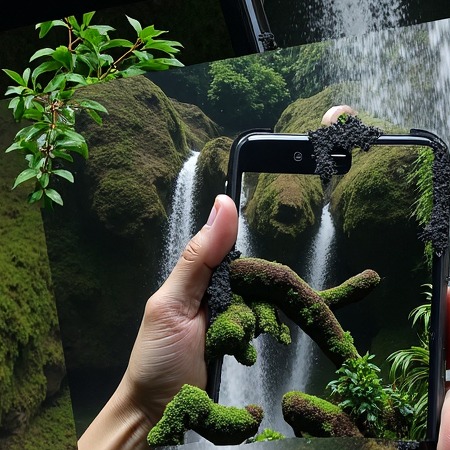}
 \caption{\small\textbf{Failure cases.} Left: recursion anchors on the window
    rather than the canvas. Right: time travel amplifies the seam
    (without, then with).\vspace{-0.3cm}}
  \label{fig:limitations}
\end{figure}

Figure~\ref{fig:ablation} compares prompting and post-processing with
cumulative sampler additions. Prompt-only and extended-prompt baselines
produce plausible scenes but lack the geometric methods' nested
structure. Post-hoc $T$ enforces repetition while carrying over scene
boundaries that the denoiser cannot adapt to the new connections.

Starting with $T$ at $\sigma=0.87$, we add warm-up, $T^\dagger$
followed by a final $T$ at $\sigma=0.5$, and in-loop SR. The
$8\times$ crops expose differences in the inner copies: with warm-up
alone, the cereal box contains a conventional illustration and the
library mirror contains shelves without another clear mirror. The
inverse/forward pair develops another nested scene; SR clarifies small
edges and details. Each column shows the cumulative effect of the
components included so far. The final column adds optional re-noising and denoising
(``time travel''), which is excluded from the main sampler.

\section{Limitations}\vspace{-0.12cm}

The fixed recursion centre may miss the intended subject. In
Figure~\ref{fig:limitations} (left), warm-up anchors the nest on a window
rather than the off-centre painter's canvas. Optional time travel can
also amplify seams instead of smoothing them, as the waterfall example
shows (right pair). More generally, finite resolution and subsequent
denoising limit geometric consistency: the continuous identities do
not guarantee exact recursion in the final raster image.

\section{Conclusion}

Braided sampling combines a frozen image model with matched geometric
maps to develop recursive scenes in source and transformed representations.
Despite finite resolution and imperfect scene consistency, the results
show how geometry can shape a composition throughout generation.
\begin{flushright}
  \small
  \textit{``I never got a pass mark in math ... mathematicians now use my prints to illustrate their books.''}\\
  ---M.~C.~Escher \citep{ernst1976}
\end{flushright}

\clearpage

\section*{Acknowledgments}
A.S. is supported by the Chaya Career Advancement Chair. This work was inspired by 3Blue1Brown’s video on YouTube: ``How (and why) to take a logarithm of an image.'' (\url{https://www.youtube.com/watch?v=ldxFjLJ3rVY})

\bibliography{refs}
\bibliographystyle{iclr2027_conference}

\clearpage
\appendix
\section{Implementation details}
\label{app:impl}

The baseline comparison in Figure~\ref{fig:ablation} uses the
displayed base prompts and a fixed initial seed within each row.
The extended-prompt baseline appends:
``In the style of M.C. Escher's Print Gallery: a recursive Droste
effect, the scene contains a smaller copy of itself spiraling inward.''
The five cumulative ablation columns each include the same central
$128\times128$-pixel crop from the $1024\times1024$ output:
columns and rows $448$ through $575$, using zero-based coordinates.
Displaying that crop at the same size as the full image gives an
$8\times$ linear enlargement. Crop locations are identical across
the compared conditions; the crops receive no separate enhancement.

\section{Sampler and hyper-parameters}
\label{app:sampler}

We build on FLUX.1-dev~\citep{flux} in half precision with its
distilled guidance at scale $3.5$ (no classifier-free guidance pass),
sampling $1024\times1024$ images with $128$ flow-matching Euler steps
and the default resolution-dependent timestep shift. On a
non-operator step we take the standard Euler update. On an operator
step we form the clean prediction
$\hat{x}_0 = x_t-\sigma_t v_\theta(x_t,t)$, decode it with the VAE,
apply the operator in pixel space, re-encode, and re-noise to the next
level, $x_{t+1}=(1-\sigma_{t+1})\hat{x}_0'+\sigma_{t+1}\epsilon$ with a
fresh $\epsilon\sim\mathcal{N}(0,I)$.

Unless stated otherwise we use inset scale
$s_{\mathrm{in}}=1/\lambda=1/4$ and $p=1$ source scale period per
output revolution.
From $\sigma_w=0.95$ to $\sigma_{\mathrm{hi}}=0.87$ (about $22$ steps)
we apply the twist-free map ($p=0$) on every step to plant the nested
copy. We then apply $T$ at $\sigma_{\mathrm{hi}}$, its generalized
inverse $\Tp$ nine steps later ($\sigma\approx0.83$), and a final $T$
at $\sigma_{\mathrm{lo}}=0.5$; the op gap sets only the $T\!\to\!\Tp$
spacing, while $\sigma_{\mathrm{lo}}$ places the last $T$. Before each
of these three window warps we upscale $\hat{x}_0$ by $\times4$ with
Real-ESRGAN~\citep{realesrgan} and area-downsample back afterwards,
which reduces resampling blur. We omit SR during warm-up because
repeated enhancement of noisy predictions can darken them. In-loop
warps use bilinear resampling. Optional time travel re-noises the
result to $\sigma_{\mathrm{tt}}=0.75$ and denoises without operators.
It may reduce joins but can also amplify seams
(Figure~\ref{fig:limitations}); it is off in the main sampler.
The nonlinear SR and denoising stages lie outside the continuous
inverse identities of Sec.~\ref{sec:pair}. A
$1024^2$ image takes about two minutes on a single $96$\,GB NVIDIA
Blackwell GPU, with peak memory about $45$\,GB when in-loop
super-resolution is active.

Tables~\ref{tab:impl-model}--\ref{tab:impl-hparams} give the exact
conformal-sampler settings; the number of window ops ($3$: $T$, $\Tp$, $T$), the final
level $\sigma_{\mathrm{lo}}=0.5$, the guidance scale $3.5$ and the
$128$ steps are held fixed. Table~\ref{tab:impl-hparams} lists the
remaining knobs with their defaults and the values we swept.

\begin{table}[h]
  \centering\small
  \vspace*{-0.15cm}\caption{Model and sampler.}
  \label{tab:impl-model}
  \begin{tabular}{@{}ll@{}}
    \toprule
    Component & Setting \\
    \midrule
    Base model & FLUX.1-dev ($12$B rectified-flow DiT), \texttt{FluxPipeline}, fp16 \\
    Guidance & distilled guidance embedding, scale $3.5$ (no CFG pass) \\
    Resolution & $1024\times1024$ (latent $128\times128\times16$, VAE factor $8$) \\
    Scheduler & FlowMatchEulerDiscrete, $128$ steps, resolution-dependent shift \\
    Non-op step & Euler flow step, $\hat{x}_0=x_t-\sigma_t v_\theta$ \\
    Op step & decode $\hat{x}_0$, warp, re-encode, re-noise to the next level \\
    Hardware & single NVIDIA Blackwell GPU ($96$\,GB); ${\approx}2$\,min/image; peak ${\approx}45$\,GB \\
    \bottomrule
  \end{tabular}
\end{table}

\begin{table}[h]
  \centering\small
  \vspace*{-0.15cm}\caption{Conformal transform $T$.}
  \label{tab:impl-transform}
  \begin{tabular}{@{}ll@{}}
    \toprule
    Item & Definition \\
    \midrule
    Sampling coordinates & $z=\exp(\beta_p\log w)$ about the focus, then source-band folding \\
    Exponents & $\beta_p=1-\mathrm{i}p\log\lambda/(2\pi)$; forward $\alpha_p=\beta_p^{-1}$ \\
    Inset scale & $s_{\mathrm{in}}=1/\lambda=1/4$ default \\
    Periods per turn & $p=1$ default; $p=0$ gives untwisted Droste repetition \\
    Generalized inverse & $\Tp=E_1B_2$ with matched bands (Sec.~\ref{sec:pair}) \\
    Resampling & bilinear \texttt{grid\_sample}, border padding, supersampling $1$ \\
    In-loop upscaling & $\times4$ Real-ESRGAN (RRDBNet x4plus, fp16), window ops only \\
    \bottomrule
  \end{tabular}
\end{table}

\begin{table}[t]
  \centering\small
  \vspace*{-0.15cm}\caption{Operator schedule over the $128$ steps ($\sigma:1\to0$).}
  \label{tab:impl-schedule}
  \begin{tabular}{@{}llll@{}}
    \toprule
    Phase & $\sigma$ range (step) & Operator & Upscaling \\
    \midrule
    Free denoising & $1.0\to0.95$ ($0$--$18$) & none & -- \\
    Warm-up & $0.95\to0.87$ ($19$--$40$) & $T_0$, every step & none \\
    $T$-cycle window & $0.87\to0.5$ & $T$ ($41$) $\to$ $\Tp$ ($50$) $\to$ $T$ at $\sigma_{\mathrm{lo}}$ ($97$) & $\times4$ SR \\
    Free denoising & $\to0$ & none & -- \\
    Time travel (optional) & post-hoc & re-noise to $\sigma_{\mathrm{tt}}{=}0.75$, denoise, no ops & -- \\
    \bottomrule
  \end{tabular}
\end{table}

\begin{table}[t]
  \centering\small
  \vspace*{-0.15cm}\caption{Swept hyper-parameters and their defaults.}
  \label{tab:impl-hparams}
  \begin{tabular}{@{}llll@{}}
    \toprule
    Name & Symbol & Default & Swept \\
    \midrule
    Inset scale & $s_{\mathrm{in}}$ & $1/4$ & $1/8,\ 1/16$ \\
    Periods per turn & $p$ & $1$ & $-1,\ 2$ \\
    Warm-up start & $\sigma_w$ & $0.95$ & off, $0.99$ \\
    First op & $\sigma_{\mathrm{hi}}$ & $0.87$ & $0.80,\ 0.92$ \\
    Op gap & steps & $9$ & $5,\ 13$ \\
    Time travel & $\sigma_{\mathrm{tt}}$ & off & $0.60,\ 0.75$ \\
    In-loop upscaling & -- & SR $\times4$ & off, bicubic \\
    \bottomrule
  \end{tabular}
\end{table}

\subsection{Transfer to PixelDiT}

To test whether the recipe depends on latent-space denoising, we rerun
it on PixelDiT-1300M (\texttt{nvidia/PixelDiT-1300M-1024px}), a
$1.3$B-parameter pixel-space diffusion transformer with no VAE. It shares
FLUX's rectified-flow contract, so the operator step carries over
unchanged, except that the VAE round trip becomes the identity and every
$T$ and $\Tp$ is a pure pixel-space warp of $\hat{x}_0$. We keep the FLUX
recipe above and match it step for step, changing only the noise
schedule: at equal $\sigma$ PixelDiT is further along, so FLUX's values
give hazy outputs. Re-matching the levels by correlating the running
estimate with the final image (warm-up $0.95\!\to\!0.96$,
$\sigma_{\mathrm{hi}}\,0.87\!\to\!0.92$,
$\sigma_{\mathrm{lo}}\,0.5\!\to\!0.69$, time travel $0.75\!\to\!0.82$)
restores contrast and keeps the recursion. A $1024^2$ image takes about
$20$ seconds on one RTX 5090 (versus $\sim$2 minutes for FLUX). The
resulting gallery is shown in Appendix~\ref{app:additional-results}.

\section{Transformation definitions}
\label{app:transformations}

We use the source scale factor $\lambda>1$ and $L=\log\lambda$
from Sec.~\ref{sec:desmit}. After centering source coordinates,
write the output-to-source sampling rule as
\[
  y(w)=s\!\left(\exp Q(w)\right).
\]
Here $Q$ is the source log-position, and a positive factor $\eta$
sets its overall image scale. Source scale repetition and angular
periodicity identify log-positions that differ by an element of
\[
  \Lambda=L\mathbb Z+2\pi i\mathbb Z.
\]
Consequently, a branch change that adds an element of $\Lambda$
does not change the sampled source colour. Define
$\beta_{k,l}=l-ikL/(2\pi)$ for integers $k,l$.

\paragraph{Conformal twist.}
For integer $p$ we use
\[
  Q(w)=\beta_{p,1}\log w+\log\eta,
  \qquad \beta_{p,1}=1-i\frac{pL}{2\pi}.
\]
A full turn changes $Q$ by $pL+2\pi i\in\Lambda$.
Thus $p$ counts signed source scale periods per output revolution.
For $w=re^{i\theta}$, a fixed source radius satisfies
$\log r+(pL/2\pi)\theta=\mathrm{const}$, giving logarithmic
spirals. Increasing $|p|$ changes their winding; the map is locally
conformal away from the origin.

\paragraph{Poles.}
For prescribed centres $c_j$ and integer pairs $(k_j,l_j)$,
\[
  Q(w)=\sum_j\beta_{k_j,l_j}\log(w-c_j)+\log\eta.
\]
Each centre contributes a local recursion structure. A turn around
$c_j$ changes $Q$ by $k_jL+2\pi i l_j\in\Lambda$, so branch
consistency holds separately at every centre. Away from the centres,
the map is locally conformal wherever
$Q'(w)=\sum_j\beta_{k_j,l_j}/(w-c_j)$ is nonzero.

\paragraph{M\"obius.}
For distinct points $a,b$, we precompose the twist with
$M(w)=(w-a)/(w-b)$. The renderer uses
\[
  Q(w)=\beta_{k,l}\bigl[\log(w-a)-\log(w-b)\bigr]
       +\log\eta.
\]
This is the two-centre case of Poles with opposite coefficients,
linking the recursion around $a$ and $b$. Different logarithm
branches differ by source periods. This coordinate construction
does not by itself specify a matched image-space inverse.

\paragraph{Square.}
For $w=x+iy$, let $\rho_\square=\max(|x|,|y|)$, and let
$\tau\in[0,1)$ be normalized arclength around the square perimeter,
starting on the positive real axis. We use
\[
  Q(w)=l\log\rho_\square+kL\tau+2\pi i l\tau+\log\eta.
\]
One perimeter traversal adds $kL+2\pi il\in\Lambda$.
The square chart replaces circular nesting with square-shaped
nesting. It is non-conformal, and its derivatives change across
the rays through the square's corners.

\paragraph{Rimrings.}
The ideal map adds a real radial term to the twist:
\[
  Q(w)=\beta_{k,l}\log w+h(|w|)+\log\eta,
  \qquad h(r)=\frac{Ar}{R_0-r}, \quad A>0.
\]
The radial term leaves angular branch consistency unchanged.
As $r\to R_0^-$, it traverses increasingly many source scale
periods, crowding repeated bands near the outer circle.
The implementation evaluates $h$ at
$\min(|w|,R_0-10^{-3})$ to regularize the divergence.
We do not apply an inverse for this transform: inversion is
numerically unstable, and the visible divisions between bands
are part of the desired appearance.

\paragraph{Rendering.}
All five maps specify output-to-source coordinates.
Writing $Q=u+i\theta$, the renderer chooses the largest-radius
scale-equivalent sample that remains inside the source canvas.
If $r_{\max}(\theta)$ is the available source radius in direction
$\theta$, including a $0.995$ safety factor, it uses
\[
  m=\left\lfloor\frac{\log r_{\max}(\theta)-u}{L}\right\rfloor,
  \qquad r_{\mathrm{src}}=\exp(u+mL).
\]
The standalone renderer samples these source positions bicubically on
a supersampled output grid and averages to the final resolution.
The in-loop implementation uses bilinear sampling
(Table~\ref{tab:impl-transform}); the coordinate rules are shared.
Replacing exact zero logarithm arguments by $10^{-9}$ avoids
evaluating the logarithm at singular centres.
The square radius is likewise floored at $10^{-9}$.
These formulas define the coordinate renderer; the matched-band
inverse identities in Sec.~\ref{sec:pair} require the corresponding
band extraction and extension construction.

\subsection{The Droste-selected generalized inverse}
\label{app:droste-inverse}

We make the continuous construction explicit for the conformal family.
Fix $p\in\mathbb Z$, $\beta_p=1-ipL/(2\pi)$,
$\alpha_p=\beta_p^{-1}$, and a logarithm branch
$\operatorname{Arg}z\in[-\pi,\pi)$.
Let $R(\theta)=r_{\max}(\theta)$ be the source radius used by the
renderer, including its safety factor. Define the source band and
its Droste folding map by
\begin{align}
 A&=\{re^{i\theta}:R(\theta)/\lambda<r\le R(\theta)\},\\
 d(z)&=\lambda^{m(z)}z,\qquad
 m(z)=\left\lfloor
 \frac{\log R(\operatorname{Arg}z)-\log|z|}{L}
 \right\rfloor .
\end{align}
Thus $d(z)\in A$, $d(q)=q$ for $q\in A$, and
$d(\lambda z)=d(z)$. The forward image operator is
\[
 F(w)=d\!\left(\exp(\beta_p\log w+\log\eta)\right),
 \qquad (Tx)(w)=x(F(w)).
\]
Here $F$ returns a source-band coordinate, not an image value.

To read this band from an Escher image, fix one output representative
for each $q\in A$:
\[
 \psi(q)=\exp\!\left(\alpha_p
       [\log q-\log\eta+n(q)L]\right),
 \qquad n(q)\in\mathbb Z.
\]
The integer $n(q)$ fixes which scale-rotated copy is read; it is chosen
from coordinates alone and held fixed for all images. In the unbounded
continuous construction one may take $n(q)=0$. For a bounded output
canvas containing a neighbourhood of the recursion centre, an explicit
available-band convention is to choose the largest integer $n(q)$ for
which $\psi(q)$ remains inside the canvas, with the rendering safety
margin. This specifies the selected Escher band $\psi(A)$.

Since $2\pi i\beta_p=pL+2\pi i$, logarithm branch changes alter
the recovered source coordinate only by an integer power of $\lambda$.
Consequently $F(\psi(q))=q$. Define
\[
 (T^\dagger y)(z)=y\!\left(\psi(d(z))\right).
\]
Operationally, this folds $z$ into the source band, finds its matched
Escher representative, and reads the colour there. In particular,
$T^\dagger y$ is scale-periodic even when $y$ violates recursion.
The extraction and extension maps are now explicit:
\[
 B_1x=x|_A,\quad (E_1b)(z)=b(d(z)),\qquad
 (B_2y)(q)=y(\psi(q)),\quad (E_2b)(w)=b(F(w)).
\]
Hence $B_1E_1=B_2E_2=\mathrm{id}$, $T=E_2B_1$, and
$T^\dagger=E_1B_2$, giving both reflexive identities in
Eq.~\ref{eq:method-round-trips}.

For $y\in\mathcal E$, $T^\dagger y$ is the unique member of
$\mathcal S$ mapped to $y$: among all preimages, it selects the Droste
extension of the recovered band. For arbitrary $y$, the selected band
also matters because its repeated copies may disagree; $TT^\dagger y$
regenerates an admissible Escher image from that band. This is an
algebraic selection, without a minimum-norm or nearest-point claim.
The construction excludes the recursion centre. Finite sampling,
interpolation, and the subsequent SR and denoising operations are
outside these exact continuous identities.

\newpage
\section{Empirical check of idempotence}
\label{app:idempotence}

\begin{figure}[h]
  \centering
  \includegraphics[ width=\textwidth]{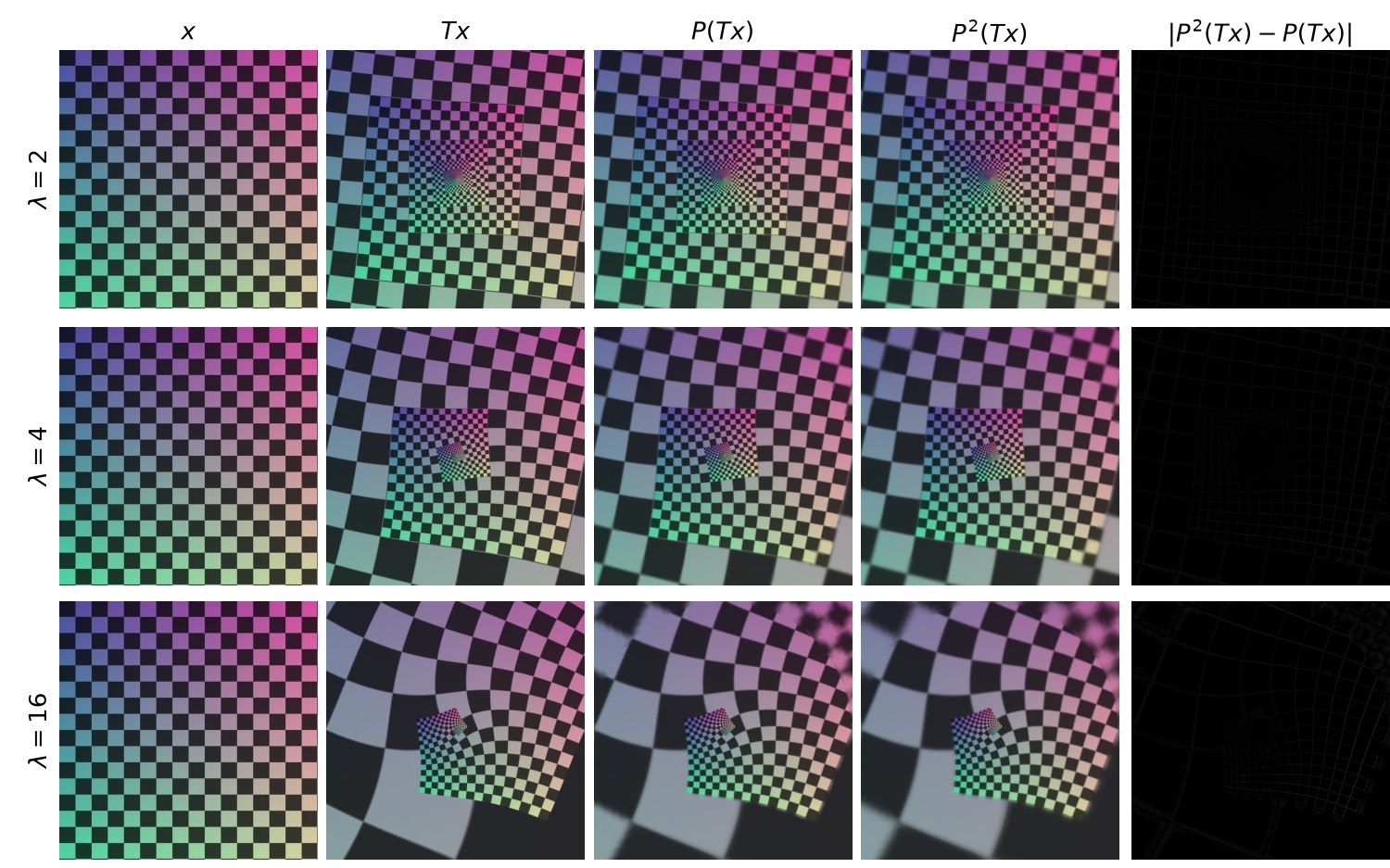}
  \vspace*{-0.15cm}\caption{\small\textbf{Checking geometric round trips.}
    From left to right: $x$, $Tx$, $P(Tx)$, $P^2(Tx)$, and the residual
    $|P^2(Tx)-P(Tx)|$, where $P=TT^\dagger$. Rows use deck ratios
    $\lambda=2,4,16$.
    The three transformed images coincide in the continuous
    construction; the rasterized comparison shows blur and
    residual differences near checkerboard edges. The measured residual
    $|P^2(Tx)-P(Tx)|$ is $0.011$, $0.012$, $0.014$ (top to bottom),
    each at the bilinear-resampling floor.}
  \label{fig:idem}
\end{figure}

We check the geometric maps separately from denoising and SR.
For $P=TT^\dagger$, an image $Tx$ is a fixed point in the continuous
construction, so $P(Tx)=P^2(Tx)=Tx$.
Figure~\ref{fig:idem} compares these images on a checkerboard and
displays the residual $|P^2(Tx)-P(Tx)|$.
The transformed panels retain the same recursive layout, while
blur and edge differences remain visible after resampling.
This is a visual check of the discretised geometric round trip;
it does not establish exact consistency of the final denoised images.

\clearpage
\section{Additional results}
\label{app:additional-results}

\subsection{Transfer to a pixel-space model}

\begin{figure}[tbp]
  \centering
  \includegraphics[width=\textwidth]{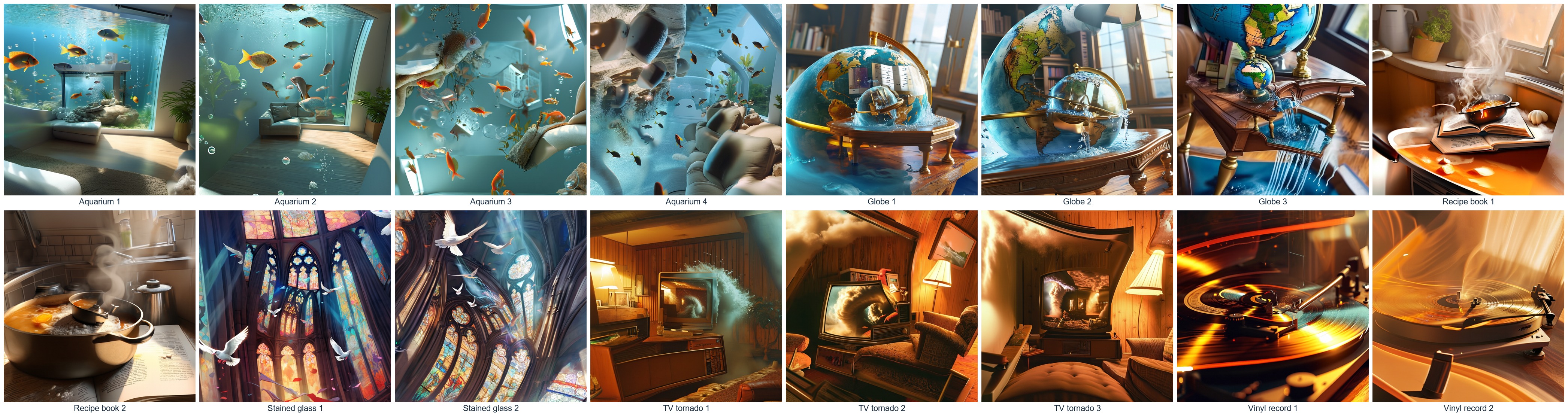}
 \caption{\small\textbf{Transfer to a pixel-space backbone.}
The conformal $T$-cycle recipe applied to PixelDiT-1300M, with the
same prompts, numerical seeds, transform, and scale as the FLUX
configuration, and adapted intervention noise levels. The warm-up
start changes from $0.95$ to $0.96$, $\sigma_{\mathrm{hi}}$ from
$0.87$ to $0.92$, and $\sigma_{\mathrm{lo}}$ from $0.5$ to $0.69$.
Where time travel is used, its noise level changes from $0.75$ to
$0.82$. Per-cell tags retain the FLUX naming convention
(e.g.\ \texttt{tt75}) and do not denote the adapted noise levels.}
  \label{fig:pixeldit}
\end{figure}

Our sampler applies its geometric operators in pixel space, so it
shouldn't be dependent on the backbone denoising in a latent space. To
test this we swap FLUX.1-dev, a latent-diffusion model, for
PixelDiT-1300M, which denoises directly in pixels, and rerun the same
recipe with only the model changed. Figure~\ref{fig:pixeldit} shows the
recursion still forms, so the technique transfers across the backbone
and across the latent-to-pixel change; the implementation is described
in Appendix~\ref{app:sampler}.

\subsection{Curated FLUX gallery}

Figure~\ref{fig:flux-gallery}, shown across the next two pages, collects
a larger set of FLUX.1-dev results, distinct from the examples in the
main text.

\begin{figure}[p]
  \centering
  \includegraphics[width=\textwidth]{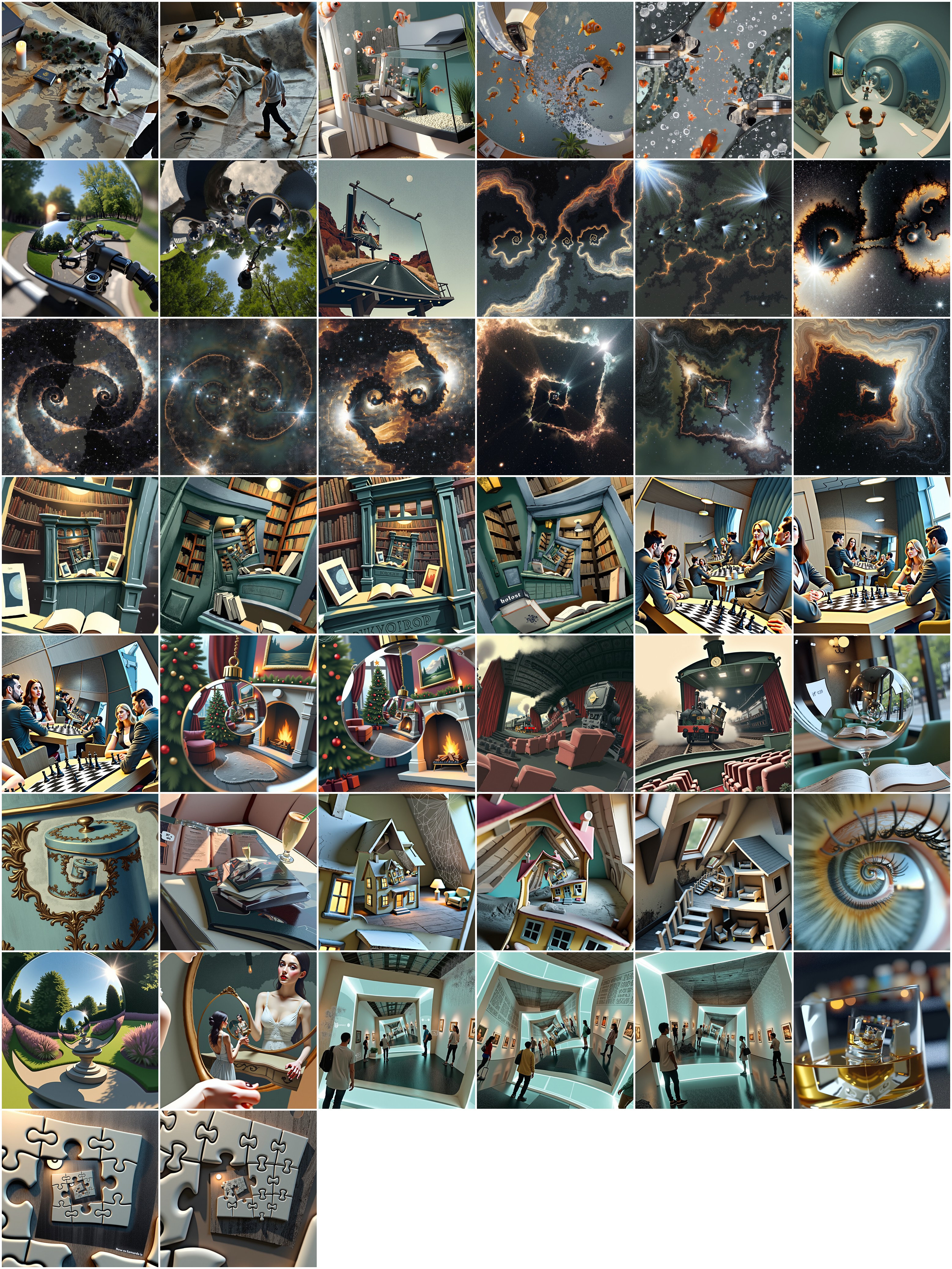}
  \caption{\small\textbf{Additional Escher-recursion results
    (FLUX.1-dev).} Curated gallery across subjects,
    transforms (conformal, square, poles, M\"obius), inset scales and
    spiral orders; 
    (Page 1 of 2.)}
  \label{fig:flux-gallery}
\end{figure}

\begin{figure}[p]
  \centering
  \ContinuedFloat
  \includegraphics[width=\textwidth]{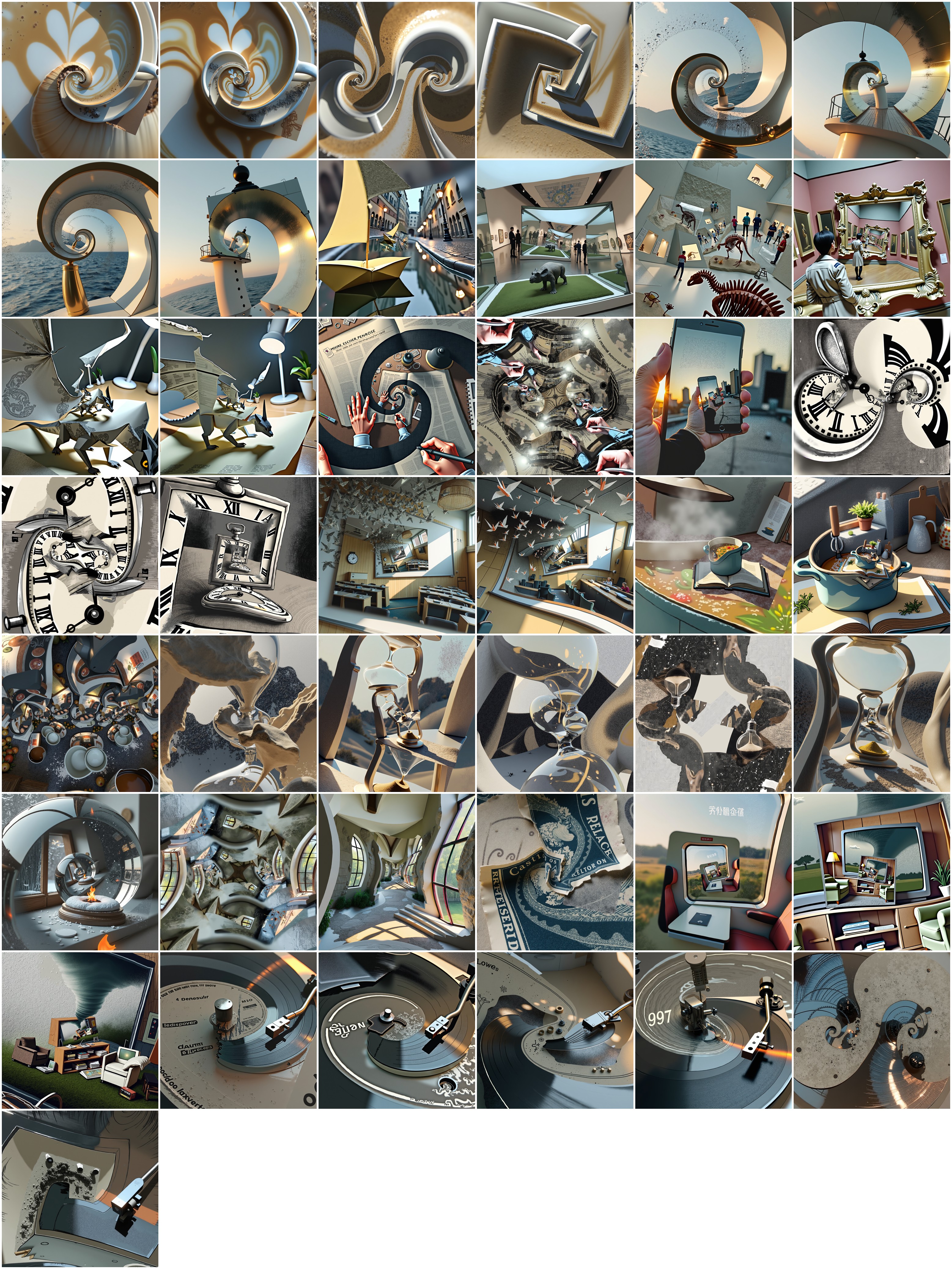}
  \caption{\small (Continued, page 2 of 2.)}
\end{figure}

\end{document}